\documentclass{article}

\usepackage{microtype}
\usepackage{graphicx}
\usepackage{subfigure}
\usepackage{booktabs} 

\usepackage{hyperref}

\usepackage[accepted]{icml2024}

\usepackage{amsmath}
\usepackage{amssymb}
\usepackage{mathtools}
\usepackage{amsthm}

\usepackage[capitalize,noabbrev]{cleveref}

\theoremstyle{plain}

\theoremstyle{definition}

\theoremstyle{remark}

\usepackage{listings}
\usepackage{color}
\usepackage{multirow}
\usepackage{bm}

\definecolor{dkgreen}{rgb}{0,0.6,0}
\definecolor{gray}{rgb}{0.5,0.5,0.5}
\definecolor{mauve}{rgb}{0.58,0,0.82}

\usepackage[textsize=tiny]{todonotes}

\newcommand{\name}{\texttt{COALA}}

\icmltitlerunning{\name{}: A Practical and Vision-Centric Federated Learning Platform}

\begin{document}

\twocolumn[
\icmltitle{\name{}: A Practical and Vision-Centric Federated Learning Platform}



\icmlsetsymbol{equal}{*}

\begin{icmlauthorlist}
\icmlauthor{Weiming Zhuang}{equal,sonyai}
\icmlauthor{Jian Xu $^\dagger$}{equal,tsinghua}
\icmlauthor{Chen Chen}{sonyai}
\icmlauthor{Jingtao Li}{sonyai}
\icmlauthor{Lingjuan Lyu}{sonyai}
\end{icmlauthorlist}


\icmlaffiliation{sonyai}{Sony AI}
\icmlaffiliation{tsinghua}{Tsinghua University}

\icmlcorrespondingauthor{Lingjuan Lyu}{lingjuan.lv@sony.com}

\icmlkeywords{Federated Learning}

\vskip 0.3in
]



\printAffiliationsAndNotice{\icmlEqualContribution} 

\begin{abstract}

We present \name{}, a vision-centric Federated Learning (FL) platform, and a suite of benchmarks for practical FL scenarios, which we categorize into three levels: task, data, and model. At the task level, \name{} extends support from simple classification to 15 computer vision tasks, including object detection, segmentation, pose estimation, and more. It also facilitates federated multiple-task learning, allowing clients to 
tackle multiple tasks simultaneously. At the data level, \name{} goes beyond supervised FL to benchmark both semi-supervised FL and unsupervised FL. It also benchmarks feature distribution shifts other than commonly considered label distribution shifts. In addition to dealing with static data, it supports federated continual learning for continuously changing data in real-world scenarios. At the model level, \name{} benchmarks FL with split models and different models in different clients. \name{} platform offers three degrees of customization for these practical FL scenarios, including configuration customization, components customization, and workflow customization. We conduct systematic benchmarking experiments for the practical FL scenarios and highlight potential opportunities for further advancements in FL. Codes are open sourced at \href{https://github.com/SonyResearch/COALA}{https://github.com/SonyResearch/COALA}.

\end{abstract}

\begin{figure*}[ht]
	\centering
	\includegraphics[width=1.98\columnwidth]{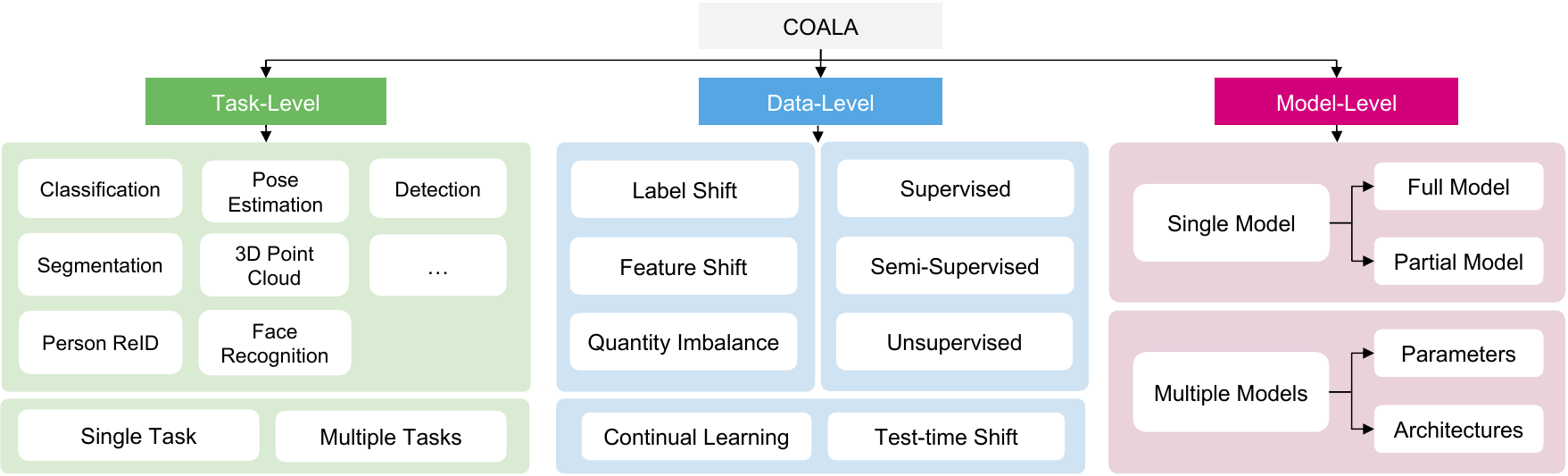}
	\caption{Illustration of three levels of practical FL scenarios supported by \name{}. At the task level, we support diverse CV tasks and training of multiple tasks in FL. At the data level, we offer out-of-the-box benchmarks for different types of data heterogeneity, various degrees of data annotation availability, and dynamic changes in data. At the model level, we extend beyond single and full model FL training to split model training and multiple model training with different architectures or parameters on clients.}
	\label{fig:functionality}
	\vspace{-2ex}
\end{figure*}

\section{Introduction}
\label{submission}

Federated learning (FL) is a distributed machine learning technique where a central server coordinates model training on decentralized clients (e.g., edge devices and institutions), which preserves data privacy by avoiding centralizing raw data from these clients \cite{fedavg, kairouz2019fl-advances-open}. It has received tremendous attention over the past few years in many application domains such as computer vision (CV) \cite{He21FedCV}, natural language processing (NLP) \cite{Wang21FLNLP,Cai23FedNLP}, recommendation systems \cite{Luo22pFLRec,Zhang23dupflrec} and information retrieval \cite{yang2023fedhap}. In particular, FL shows a great potential in computer vision applications as they are often associated with critical privacy issues due to sensitive information in images (e.g., facial recognition \cite{NiuD22fedface} and medical imaging \cite{nguyen2022fedhealth}). Therefore, it is of significant interest and importance to exploit the potential of FL across broader CV tasks.

To facilitate the fast development of FL, many FL benchmarks and libraries have been developed, the majority of which focus on addressing the data heterogeneity and scalability challenges. Particularly, several benchmarks \cite{leaf, hu2020oarf, lai2022fedscale, he2020fedml} provide datasets to simulate data heterogeneity in FL, where the data distributions among clients are different. Besides, considerable efforts of FL frameworks or libraries \cite{lai2022fedscale, garcia2022flute, huba2022papaya, beutel2020flower, Bonawitz2019FL-sys-scale, Zhang22Felicitas} aim to support the simulation or real-world deployment of FL at scale. However, 
these benchmarks and libraries cannot well support the latest development of FL in CV applications in the following three aspects:

\textbf{Task level.} Most research works are limited to the simple image classification task \cite{fedavg}. Other more challenging but practical CV tasks such as object detection had been mostly overlooked. Although several benchmarks support several other CV tasks, they are either task-specific \cite{zhuang2020fedreid} or consider only one task in FL at a time \cite{He21FedCV, hu2020oarf}. Recent work \cite{zhuang2023mas} has demonstrated the potential of training multiple CV tasks together with a significantly improved trade-off between performance and efficiency. 

\textbf{Data level.}  The majority of works focus only on supervised learning, where all the training samples in clients contain labels. Recently, semi-supervised or unsupervised FL has emerged as an important topic as it is hard to obtain data labels on clients \cite{Diao22SemiFL,Zhuang22FedEMA}. In addition, the existing libraries mostly consider skewed label distribution in data heterogeneity, while the feature shifts where the data in each client is from different domains are often overlooked. Moreover, static data distribution among clients is usually assumed, which however may not hold in realistic scenarios, e.g., the distribution of images captured from street cameras in smart cities could evolve over time.

\textbf{Model level.} Most works predominantly consider only a single model with complete model architecture trained in clients. However, many works have demonstrated the feasibility of spitting a model and offloading a part of the model to the server to reduce computation on clients \cite{thapa2022splitfed,Li23MocoSFL}. Besides, recent studies consider more practical scenarios of training distinct models in different FL clients \cite{Diao0T21HeteroFL,qin2023fedapen}. Moreover, adopting foundation models in FL, especially by parameter-efficient fine tuning (PEFT), also draws increased attention \cite{zhuang23FMFL,Zhang23FedPETuning,Herbert24SurveyFLFM}. Supporting and benchmarking these scenarios are vital for the further development of CV and FL in heterogeneous systems.

\textbf{Contribution}. In this work, we propose a new vision-centric FL platform, named \name{}, together with a suite of FL benchmarks to bridge gaps in these three levels.
\name{} offers extensive benchmarks for diverse vision tasks and various new learning paradigms in FL, which,
to the best of our knowledge, covers the most comprehensive and up-to-date research topics in current FL studies.
At the task level, we extend 
to a broader spectrum of 15 CV tasks, including classification, object detection, segmentation, pose estimation, face recognition, and more (refer to Table~\ref{tab:overview}). Additionally, we facilitate federated multiple-task learning, enabling clients to simultaneously train on more than one task. At the data level, \name{} expands upon supervised FL and label distribution shift in data heterogeneity: 
it can support semi-supervised FL, unsupervised FL, and multi-domain FL with feature distribution shifts among local training data. Furthermore, \name{} caters to federated continual learning, accounting for continuously changing data in clients in practical scenarios. At the model level, \name{} supports traditional training of a single full model, computationally efficient split learning, and federated multiple-model training where clients can train multiple models with varying parameters and architectures.

\name{} seamlessly supports new FL scenarios with three degrees of customization: configuration customization, components customization, and workflow customization (Section \ref{sec:platform}). It provides automated benchmarks and evaluations for a suite of practical FL scenarios in the task, data, and model levels (Section \ref{sec:benchmark}). Our systematic experiments showcase the capability of \name{} to deliver comprehensive benchmarking 
across various FL scenarios. These results demonstrate the promising potential of \name{} while also showing the opportunities for further enhancements in the emerging domains of CV and FL.
\vspace{-1ex}


\section{Related Work}

\textbf{Prior FL platforms}. In the past few years, there have been a bunch of open-source platforms and tools being developed to facilitate algorithm evaluations and practical applications. For example, the FATE \cite{fate}, FederatedScope \cite{Xie23FederatedScope}, FedML \cite{he2020fedml}, OpenFed \cite{chen2023openfed}, just name a few, are among the most famous frameworks released by industrial companies, while LEAF \cite{leaf}, Flower \cite{beutel2020flower}, FedScale \cite{lai2022fedscale}, EasyFL \cite{Zhuang22EasyFL}, FLUTE \cite{garcia2022flute}, FedLab \cite{Zeng23FedLab} are mainly 
contributed by academia. Other frequently mentioned platforms in the literature include TensorFlow Federated \cite{tff}, PaddleFL \cite{paddlefl-github}, PySyft \cite{pysyft} and FedVision \cite{liu2020fedvision}. We also notice that some recent works, such as Felicitas \cite{Zhang22Felicitas}, PAPAYA \cite{huba2022papaya}, Flint \cite{wang2023flint} and FS-Real \cite{chen2023FS-REAL}, pay more attention to the real-world constraints for more realistic device-cloud collaborative FL. It is worth noting that some frameworks mainly focus on supporting flexible simulations for research purposes 
only, including FLUTE \cite{garcia2022flute} that aims at rapid prototyping of new algorithms at scale and FLGo \cite{wang2023flgo} that customizes FL tasks with shareable components and plugins. While each platform has its own specialties and advantages, there is still no platform that specializes and enables comprehensive vision tasks under practical FL settings (continual learning; multi-domain data, etc). 
By contrast, our proposed \name{} is the first vision-centric FL platform that can be used for both distributed training simulation and realistic cross-device applications with high flexibility in customization.

\textbf{Existing FL benchmarks}. LEAF \cite{leaf} is the first FL benchmark that provides some datasets with statistical heterogeneity. However, the supported vision tasks and datasets are very limited, and only supervised learning is considered. FedReID \cite{zhuang2020fedreid} improved the federated person re-identification via benchmark analysis. An experimental study of representative FL algorithms for image classification on non-IID data silos was provided in \cite{li2022federated}. OARF \cite{hu2022oarf} provided a benchmark suite that is diverse in data size, label distribution, feature distribution, and learning task complexity. FedScale \cite{lai2022fedscale} provided the natural partitions of real-world datasets with real client-data mapping to better simulate large-scale FL settings. The most related work is FedCV \cite{He21FedCV}, which is the first work that evaluates representative vision tasks in FL settings, including classification, object detection, and segmentation. However, with the growing of FL in real-world applications, those benchmarks are not comprehensive enough to assess the effectiveness of FL algorithms on diverse vision tasks under various real scenarios. We also notice some benchmarks specialized on other tasks and applications, including audio \cite{zhang2023fedaudio}, NLP \cite{Lin22FedNLP}, multi-modal learning \cite{Feng23FedMultimodal}, IoT \cite{alam2023fedaiot}, and model personalization \cite{chen2022pflbench}, etc.
\vspace{-1ex}

\section{\name{} Benchmark: Practical FL Scenarios}
\label{sec:benchmark}
In this section, we first introduce the basic FL protocol, followed by the practical FL scenarios supported by our \name{} benchmark from task level, data level, and model level, as summarized in Figure \ref{fig:functionality}.
\subsection{Basic FL Protocol}
We consider a typical FL setup with $m$ clients that collaboratively train a global model with parameters $\bm{w}$ under the coordination of a server, which could be formalized by:
\begin{equation}\label{eq:objective}
	\min_{\bm{w}}{L({\bm{w}})} = \sum_{i=1}^{m}\frac{|\mathcal{D}_i|}{|\mathcal{D}|} \underbrace{ \mathbb{E}_{(x_i,y_i)\sim \mathcal{D}_i}\left[l(f(x_i;\bm{w}), y_i)\right]}_{:= L_i(\bm{w})},
\end{equation}
where $\mathcal{D}_i$ is the local training data in the client following an underlying distribution $\mathcal{P}_i(X,Y)$ on $\mathcal{X}_i \times \mathcal{Y}_i$, where $\mathcal{X}_i$ is the input space and  $\mathcal{Y}_i$ is the label space. $\mathcal{D}$ denotes the collection of all training data among clients. $f(x_i;\bm{w})$ and $l(f(x_i;\bm{w}), y_i)$ denote the model output and loss function, respectively, given parameters $\bm{w}$ and a data point $(x_i,y_i)$. The objective function $l(\cdot,\cdot)$ varies based on specific tasks, such as cross-entropy loss for classification and Mean Square Error (MSE) for regression. The global objective $L(\bm{w})$ can be regarded as a weighted average of local objectives. FedAvg~\cite{fedavg} is the de facto algorithm in FL, with advanced methods following this basic protocol while introducing new strategies in client training or server aggregation \cite{reddi2020fedyogi}.

\subsection{Task Level: Diverse CV Tasks and Multiple Tasks}

\name{} provides out-of-the-box support for 15 vision tasks, including classification, detection, segmentation, pose estimation, face recognition, person re-identification (ReID), 3D point cloud, and more. A comprehensive overview of the tasks, datasets, and models is presented in Table \ref{tab:overview} in the Appendix. To the best of our knowledge, \name{} is the most comprehensive FL platform for vision tasks and it is also highly customizable and easy to extend to new tasks. Beyond single-task training, \name{} also facilitates the concurrent training of multiple tasks.

\textbf{Federated Image Classification}. Classification is the most common task in FL \cite{fedavg,fedprox}, where the label space could be expressed by $\mathcal{Y} = \{1,.., C\}$ and $C$ is the total number of categories. \name{} offers seven datasets with varying scales and difficulty levels for classification. It enables the simulation of diverse non-IID data types, varying degrees of data annotation availability, and dynamic changes in data, detailed in Section \ref{sec:data-level}.

\textbf{Federated Object Detection}. 
Object detection is a pivotal vision task in practical applications such as autonomous driving systems. However, federated object detection has received much less attention than classification \cite{liu2020fedvision, kim2023navigating}. It follows the basic FL protocol, but with more complex data annotations as each image often contains multiple objects, each annotated with category labels and bounding box positions. As a result, it is much more challenging to simulate label distribution shifts for the detection task. \name{} introduces two different non-IID simulations (detailed in Appendix~\ref{app:visual}) on BDD100K dataset \cite{yu2020bdd100k}, which is a representative driving dataset for object detection. In addition, our platform supports both single-stage detectors like YOLO \cite{redmon2016yolo} and two-stage detectors like Faster RCNN \cite{ren2015faster}.

\textbf{Federated Semantic Segmentation}. Segmentation is another important vision task widely used in autonomous driving and medical image analysis. Training samples in segmentation are annotated in a pixel-wise manner, making it more challenging to obtain sufficient labeled data in a single client. This difficulty raises the need for collaborative learning using FL. Meanwhile, the non-IID data also brings new challenges as different clients may have different objects and there may exist foreground-background inconsistency \cite{miao2023fedseg}. For this task, we integrate the Pascal VOC \cite{everingham2010pascal} with label shift for simulation and 
DeepLab models \cite{chen2018deeplab} for benchmarking.

\textbf{Federated Pose Estimation}. Pose estimation aims to detect the position and orientation of a person or an object \cite{toshev2014deeppose}. To the best of our knowledge, no prior studies in FL investigate this task and provide empirical results. To bridge this gap, we integrate the popular dataset MPII Human Pose \cite{andriluka14mpii} with IID data partition to provide an initial exploration in FL. 

\textbf{Federated Face Recognition and Person ReID.} The widespread use of face recognition (e.g., airport check-in and mobile Face ID) has sparked concerns about individual privacy. FL can provide privacy-aware training for face recognition models without access to private face images from clients \cite{Meng22Face,NiuD22fedface,Liu22FedFR,zhuang2022fedfr}. Person re-identification (ReID) \cite{zhuang2020fedreid,zhuang2021joint,zhuang2023optimizing} aims at matching a person's identity across different cameras or locations. \name{} integrates four datasets for face recognition and eight datasets for person ReID from independent sources to mimic the decentralized data with feature shift.

\textbf{Federated Multiple-Task Learning}. In addition to training a single vision task, federated multiple-task learning emerges as a new FL scenario where each client trains multiple tasks efficiently and simultaneously under resource constraints \cite{bhuyan2022multi}. These tasks can be trained separately with multiple models or jointly by adopting multi-task learning (MTL) with a shared backbone encoder and multiple task-specific decoders \cite{zhuang2023mas}. We integrate the widely used Taskonomy dataset \cite{zamir2018taskonomy} with each client containing data from one building to simulate the statistical heterogeneity and quantity imbalance to evaluate the potential of federated multiple-task learning.

\subsection{Data Level: Realistic Data Patterns}
\label{sec:data-level}

\name{} provides flexible support for different data patterns arising in FL, including the distribution shift (label shift, feature shift, test-time shift), quantity imbalance, continual learning, and different availability of data annotations (supervised, semi-supervised, unsupervised). We provide benchmarks and simulations for these realistic data patterns to evaluate FL algorithms.

\textbf{Training Data from Multiple Domains.} To tackle the data heterogeneity issue in FL, most studies use the single domain datasets with \textit{label shift} (e.g., shared input space $\mathcal{X}_i=\mathcal{X}, \forall i \in [m]$) for evaluating the performance of either global model or local personalized models. However, in real-world applications, due to the diverse environments and independent data collection, data samples could originate from different domains, resulting in \textit{feature shifts} (different feature distributions) among clients (i.e., $\mathcal{X}_i \ne \mathcal{X}_j, i,j \in [m]$). The most recent works study FL with feature distribution shifts \cite{zhuang2023normalization,zhuang2024fedwon}, using datasets such as Digits-Five, Office-Caltech, and DomainNet \cite{li2020fedbn}. \name{} provides benchmarks using these datasets for the classification task. It also supports other tasks like person ReID with feature shifts.
The BDD100K data collected from different weather, scene, and time-of-day attributes can also be considered as feature shifts.


\textbf{Federated Semi-supervised Learning.} As annotating all training data is time-consuming and expensive, it is usually impractical to assume that all the clients could have fully labeled training sets. Most clients are likely to have only a small portion of data labeled, with the remaining data being unlabeled; in some cases, all clients might be unlabeled while the central server has labeled data \cite{zhang2021improving,Liang22RSCFed,Diao22SemiFL}. These scenarios are referred to as label-in-client and label-in-server situations in the literature \cite{Jeong21FedMatch}. \name{} provides benchmarks for both scenarios. The goal of semi-supervised FL is to leverage the unlabeled data to train a better model without violating the privacy of local data.

\textbf{Federated Unsupervised Learning.} In an extreme case where all the clients only have unlabeled data, unsupervised learning can be adopted into the FL. \name{} supports both federated self-supervised learning to learn generic visual representations \cite{Zhuang22FedEMA} and federated unsupervised learning for specific tasks like person ReID. Besides, we provide a benchmark using one popular self-supervised learning method called BYOL \cite{grill2020bootstrap}.

\textbf{Federated Continual Learning.} Conventional FL assumes data in clients remains static, yet in reality, data can be dynamically changing over time. Federated continual learning emerges as a solution to continuously update the model with the new data while preventing forgetting the old knowledge \cite{zhang2023target}. \name{} provides a benchmark for federated class-continual learning, enabling clients to continuously collect and learn new data classes. This introduces variability in the label space $\mathcal{Y}^t$ across different time slots $t$, with a typical setting where $\mathcal{Y}^t \cap \mathcal{Y}^{t'}=\varnothing$ for $t \ne t'$.

\textbf{Test-time Distribution Shift.} Addressing test-time distribution shifts is an emerging challenge in FL, where the data distribution could differ in training and testing. Such shifts could be label shifts, feature shifts, or covariate shifts \cite{tan2023taming,tan2024heterogeneity}. \name{} provides benchmarks for test-time distribution shift in FL. The challenge is how to utilize the diverse data in FL to learn either a shift-resilient model \cite{tan2023taming} or better adaptation strategies \cite{bao2023adaptive}.

\subsection{Model Level: Model Configuration}
Rather than training a single full model as in typical FL, \name{} provides comprehensive support for different model configurations. This includes splitting the full model or training different models for different clients. 

\textbf{Single Model.} The majority of FL studies consider training a single model, where the model architecture is the same among clients and the participating clients start training from the same model parameters. We term it as training the \textit{full model}. In contrast, some FL works only transmit and communicate a \textit{partial model} among clients and the server. For example, several studies only communicate the backbone while keeping the classifier locally in clients \cite{zhuang2020fedreid,dong2022spherefed}. 
\name{} provides benchmarks for both the full model and partial model. To further support the latest development of foundation models, we further support federated parameter-efficient fine-tuning (PEFT) \cite{Hu22LoRA} of foundation models.

\textbf{Multiple Models.} In addition to a single model, \name{} also supports benchmarking for the configuration of multiple models. We define multiple models as either the model architectures in clients are different or the model parameters with which the clients start training 
are different. These settings reflect the practical FL scenarios: different model architectures could address system heterogeneity in FL where clients could have varied resources \cite{Diao0T21HeteroFL}; different model parameters in clustered FL and personalized FL for varied data distribution in clients \cite{li2019fedmd}.

\begin{figure}[t]
\centering
\includegraphics[width=\columnwidth]{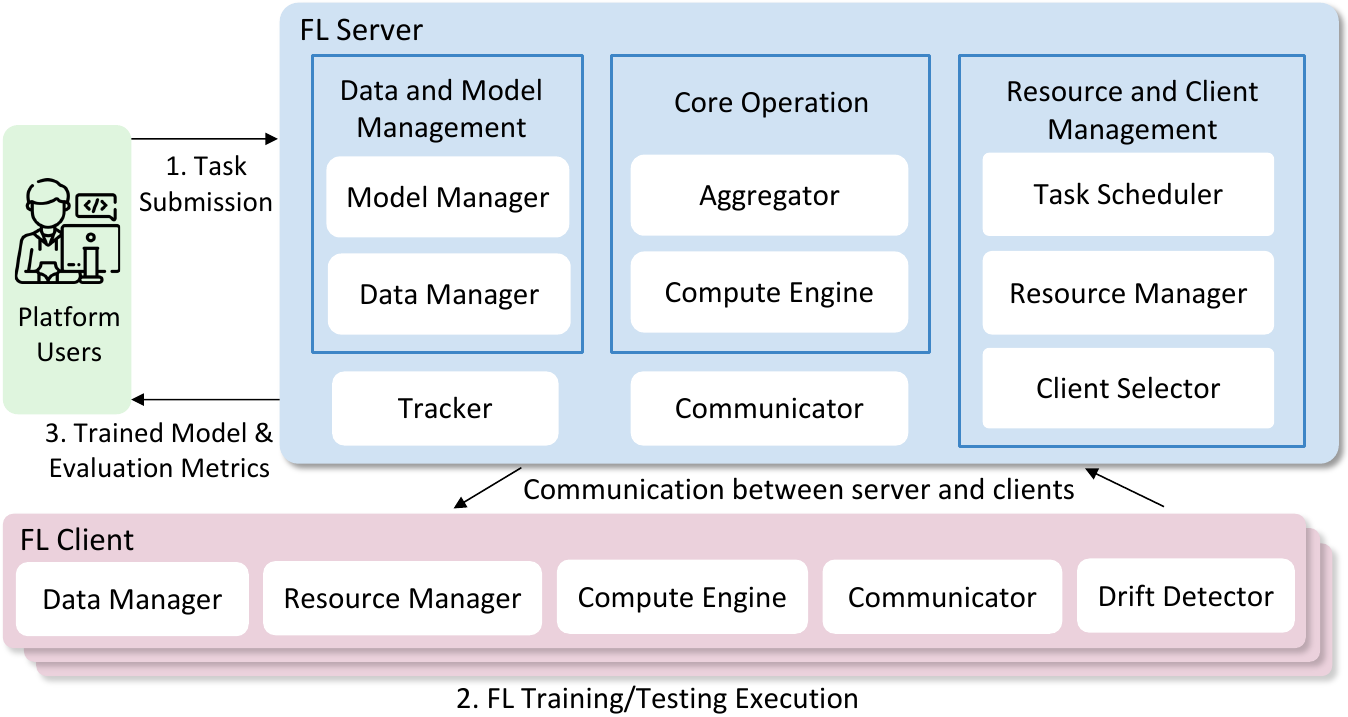}
\vspace{-3ex}
\caption{Illustration of \name{} platform that enables automated benchmarking for practical FL scenarios.}
\label{fig:system-design}
\end{figure}

\section{\name{} Platform: System Design} 
\label{sec:platform}

Existing FL platforms cannot adequately support diverse and practical FL scenarios as discussed in Section \ref{sec:benchmark}. To bridge this gap, we present \name{}, a new FL platform designed for the latest advancements in the field. \name{} automates FL training and evaluation and is scalable and highly customizable. Figure \ref{fig:system-design} depicts the overall architecture of \name{}. We discuss the system workflow and three degrees of customizations it supports in this section and provide details of the system components in Appendix \ref{app:component}.

\subsection{System Workflow}

\name{} platform streamlines and automates the FL workflow, providing a seamless process from training task initiation to output delivery.

\textit{Task Submission and Initialization}: Users commence the FL process by submitting configurations and customized components to \name{} platform. Then, the task scheduler works with the resource manager to allocate resources for the server to start a new FL training task. 

\textit{FL Training and Testing Execution}: At the start of training each task, the data manager and model manager load the dataset and model for training, respectively. The server then selects FL clients based on the resource availability of the clients. Then, the system executes the standard FL life cycle: 1) the server distributes the task (configurations and models) to the clients via communicator; 2) the client executes the training/testing and then uploads the results and trained model parameters to the server; 3) the aggregator in the server aggregates these models and obtains a new model for the next round of training.

\textit{Output Delivery}: The server tracker collects and consolidates the evaluation metrics from the server and clients. The evaluation metrics contain both system metrics (e.g., computation time, communication time, memory consumption, etc.) and algorithmic metrics (e.g., accuracy and loss). These evaluation metrics together with the trained model are delivered to the users at the end of training.

\subsection{Three Degrees of Customization}
The \name{} platform is highly customizable to support a wide range of practical FL scenarios. Users can customize their FL applications and algorithms in three different degrees: configuration degree, component degree, and workflow degree. We summarize the procedure of our customization in Appendix~\ref{app:customize}.

\textit{Configuration Customization:} The first and easiest customization is through configurations. \name{} platform loads a configuration file to set up the initial FL settings. Users can easily modify configurations to use out-of-the-box datasets, models, algorithms, and FL settings (e.g., number of participating clients and number of local epochs). At the data level, users can customize the dataset for training and the way to partition the dataset (e.g., label shifts and domain shifts). At the model level, \name{} allows them to customize the model for training with the full model. An example of the configuration is provided in Appendix \ref{app:customize}.

\textit{Component Customization:} \name{} extends to component customization, supporting users to customize system components and load them to the platform as plugins. This includes customization of data, model, server and client executions, which are not provided by the platform. Leveraging our modular design, users can inherit existing implementations and easily register customized components into the platform via APIs. List in Appendix~\ref{app:customize} offers an illustrative example of customizing client implementation, including training execution, testing execution, and content uploads. This component-level customization facilitates many FL scenarios in Section \ref{sec:benchmark}, such as training diverse models at the model level; Users can personalize server distribution and customize aggregation methods like adopting knowledge distillation \cite{lin2020ensemble} or clustering \cite{ghosh2020efficient}. \name{} then executes the standard FL workflow with these plugins.

\begin{figure*}[t]
	\centering
	\subfigure[Digits-Five]{
		\includegraphics[width=0.6\columnwidth]{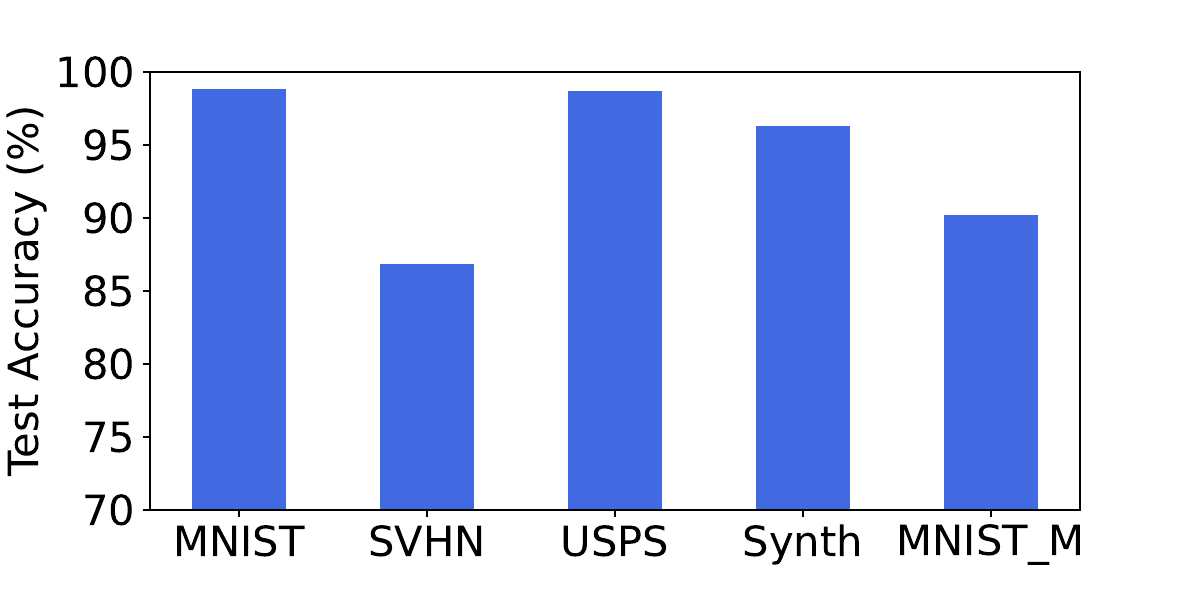}
	}
		\quad
	\subfigure[Office-Caltech]{
		\includegraphics[width=0.6\columnwidth]{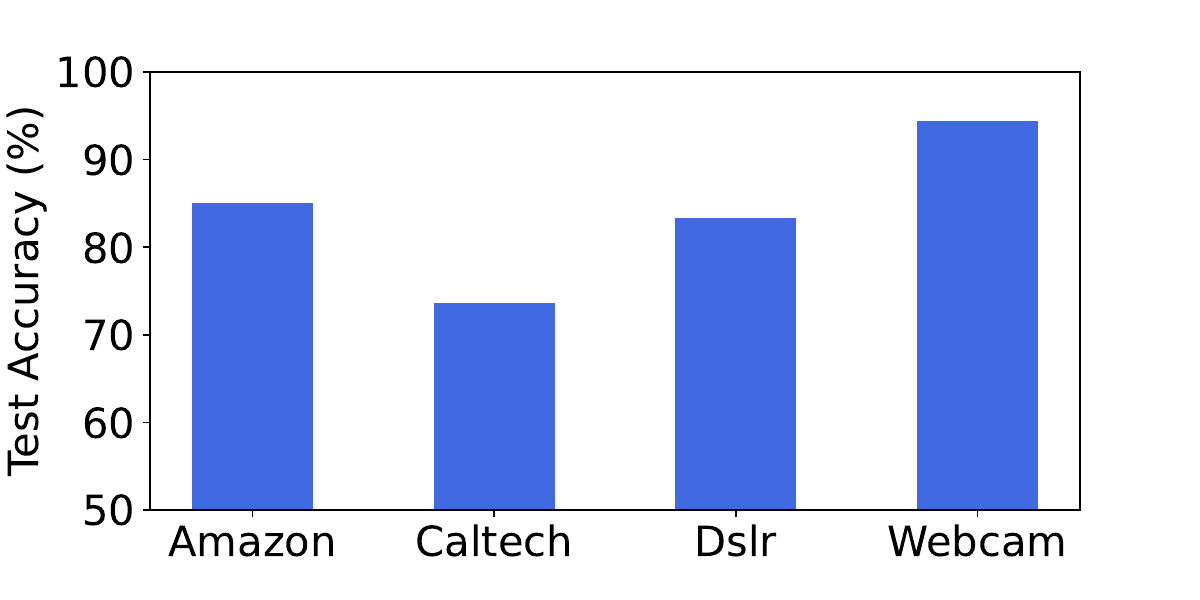}
	}
         \quad
	\subfigure[DomainNet]{
		\includegraphics[width=0.6\columnwidth]{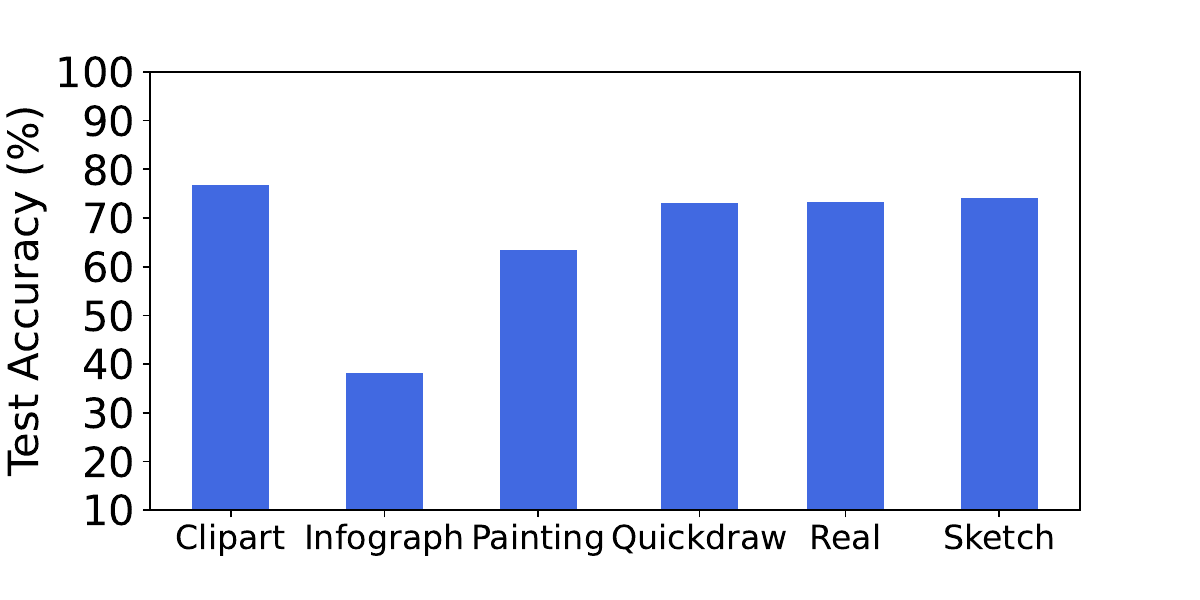}
	}
	    \vspace{-2ex}
	\caption{Domain-wise test accuracy of the global model. The same model does not perform equivalently well on different domains.}
	\label{fig:multi-domain}
\end{figure*}

\textit{Workflow Customization:} In addition, \name{} allows users to customize FL workflow (i.e., execution logic in both the FL server and the FL client), other than standard FL workflow. \name{} platform wraps the server execution logic inside a function in the server. Users can customize the server workflow by inheriting and extending to a new server execution. At the same time, they can still customize components or reuse the out-of-the-box components in the platform. For instance, in the semi-supervised FL scenario, users can design a customized FL workflow where the server initially trains with publicly available data, distributes to clients for unsupervised training, fine-tunes the aggregated model with additional public data post-aggregation, and subsequently distributes the refined model to clients for the next round of training. Such customization empowers users to tailor FL workflows to their unique requirements, opening more possibilities for new application and algorithm development.

\section{Benchmark Experiments}
\label{sec:experiments}

In this section, we present representative benchmark results to show how \name{} facilitates effective implementation and benchmarking of practical FL scenarios. We conduct the benchmark mostly using FedAvg and more details of experimental settings are provided in Appendix~\ref{app:exp_setup}.

\subsection{Task Level Benchmark}
\label{sec:exp-supervised}

\textbf{Image Classification:} 
We benchmark classification with the standard CIFAR-10/100 datasets and partition it into 100 clients with label distribution shifts using the Dirichlet distribution with the concentration parameter $\alpha=0.5$ (Dir(0.5)). Table~\ref{tab:classification} shows that label distribution shift affects the model performance, which aligns with other FL studies.

\begin{table}[t]
	\small
	\centering
	\caption{Benchmark of federated image classification task.}
		\renewcommand\arraystretch{0.9}
	\begin{tabular}{ l c c c}
		\toprule[1pt]
		\textbf{Datasets} & \textbf{\# Client} &  \textbf{Heterogeneit}y  &  \textbf{Acc. (\%)} \\
		\midrule
            CIFAR10 & 100 & IID & 90.90
		\\
		CIFAR10 & 100 & Dir(0.5) & 84.08
		\\
            CIFAR100 & 100 & IID & 63.52
		\\
		CIFAR100 & 100 & Dir(0.5) & 56.80
		\\
            \midrule
		Digits-5 & 5 & Feature shift & 94.14
		\\
		Office-Caltech & 4 & Feature shift & 84.09
		\\
		DomainNet & 6 & Feature shift & 66.47
		\\
		\bottomrule[1pt]
		\label{tab:classification}
	\end{tabular}
		\vspace{-4ex}
\end{table}

\begin{table}[t]
		\small
	\centering
	\caption{Benchmark of federated object detection on BDD100K.}
	\renewcommand\arraystretch{0.9}
	\begin{tabular}{ l l c c c}
		\toprule[1pt]
		\textbf{\# client} &  \textbf{Attributes} &  \textbf{Precision} & \textbf{Recall} & \textbf{mAP@0.5} \\
		\midrule
		10 & IID & 61.06 & 32.68  & 34.45  
		\\
            10 & Dir(0.5)& 59.41 & 32.71  & 33.43  
		\\
		10 & H-Dir(0.5) & 58.91 & 29.89  & 31.36  
		\\
		\midrule
		100 & IID & 58.84 & 29.24 & 30.71 
		\\
		100 & Dir(0.5)& 58.64 & 30.61 & 32.07
		\\
		100 & H-Dir(0.5)& 58.75 & 29.82 & 31.42  
		\\
		\bottomrule[1pt]
		\label{tab:detection}
	\end{tabular}
		\vspace{-4ex}
\end{table}

\textbf{Object Detection:} We provide the benchmark for federated object detection using the BDD100K dataset for the autonomous driving application. We simulate data heterogeneity in two ways: 1) a Dirichlet distribution-based partition strategy on weather attributes (\textit{Dir}); 2) a hierarchical strategy by repeating Dirichlet distribution-based partition on all three attributes (weather, scene, time-of-day) (\textit{H-Dir}). Both result in feature shifts and data quantity imbalance among clients, as visualized in Appendix~\ref{app:visual}. The model we use is the YOLO-V5\footnote{https://github.com/ultralytics/yolov5} model.
Table~\ref{tab:detection} illustrates the effective training of object detection models in FL.  The \textit{H-Dir} strategy causes greater divergence in data distribution than the \textit{Dir} strategy, thus resulting in lower performance. 
It is also interesting to see that the feature shift along with data quantity imbalance may not cause a significant performance drop, and could even achieve slightly better performance. This could be because the randomly selected clients contain more data in \textit{Dir} simulation than the IID setting, as data quantity is highly skewed (we randomly select 4 out of 10 clients and 12 out of 100 clients each for these experiments). Deeper investigations could present potential opportunities for future research.



\begin{figure}[t]
	\centering
	\includegraphics[width=0.85\columnwidth]{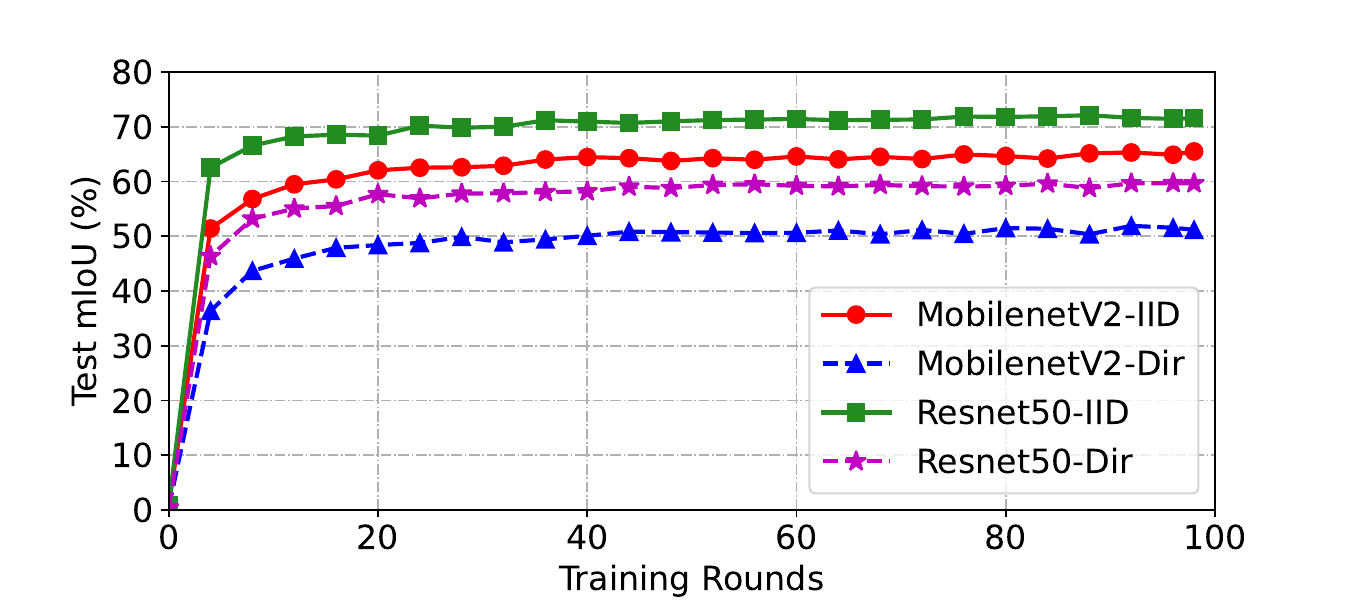}
 \vspace{-2ex}
	\caption{Benchmark of federated semantic segmentation on VOC.}
	\label{fig:seg}
		\vspace{-4ex}
\end{figure}

\textbf{Semantic Segmentation:} We train the semantic segmentation model DeepLabV3 on the PASCAL VOC with pre-trained MobileNet-V2 and ResNet-50 as backbones, respectively. We simulate 10 FL clients and randomly select 4 in each training round.
We use the first category of semantic objects in each image as the label for the label distribution shifts simulation. Figure~\ref{fig:seg} indicates that the label shift problem severely degrades the performance using both the MobileNet-V2 and ResNet-50. 
These benchmark results suggest that non-IID is also a significant issue in segmentation, which could be opportunities for robust algorithms and solutions to address this challenge.

\textbf{Pose Estimation:} \name{} provides benchmark results for pose estimation with the MPII dataset on three different sizes of ResNet backbones.
Since the human has multiple joints to be estimated, we focus on the Mean performance for the Percentage of Correct Keypoints (PCK@0.5). Table~\ref{tab:pose} demonstrates that the pose estimation model achieves competitive performance in FL. However, it is interesting to find that a larger backbone does not have a significant impact on performance.

\begin{table}[t]
	\small
	\centering
	\caption{Benchmark of federated pose estimation on MPII dataset.}
		\renewcommand\arraystretch{0.9}
	\begin{tabular}{ l c c c}
		\toprule[1pt]
		\textbf{Backbone} &  ResNet-34 &  ResNet-50 & ResNet-101\\
		\midrule
		 Head   & 96.08 & 96.39 & 96.17
		 \\
		 Shoulder  & 94.56 & 94.48 & 94.65
		 \\
		 Elbow  & 86.84 & 87.66 & 87.63
		 \\
		 Wrist  & 80.83 & 82.15 & 82.75
		 \\
		 Hip  & 87.46 & 86.95 & 87.66
		 \\
		 Knee  & 81.58 & 82.03 & 82.39
		 \\
		 Ankle  & 77.49 & 77.07 & 77.33
		 \\
		 \midrule
		 \textbf{Mean}  & 87.04 &  87.30 & 87.57
		 \\
		\bottomrule[1pt]
		\label{tab:pose}
	\end{tabular}
		\vspace{-5ex}
\end{table}

\begin{figure}[t]
	\centering
	\includegraphics[width=0.81\columnwidth]{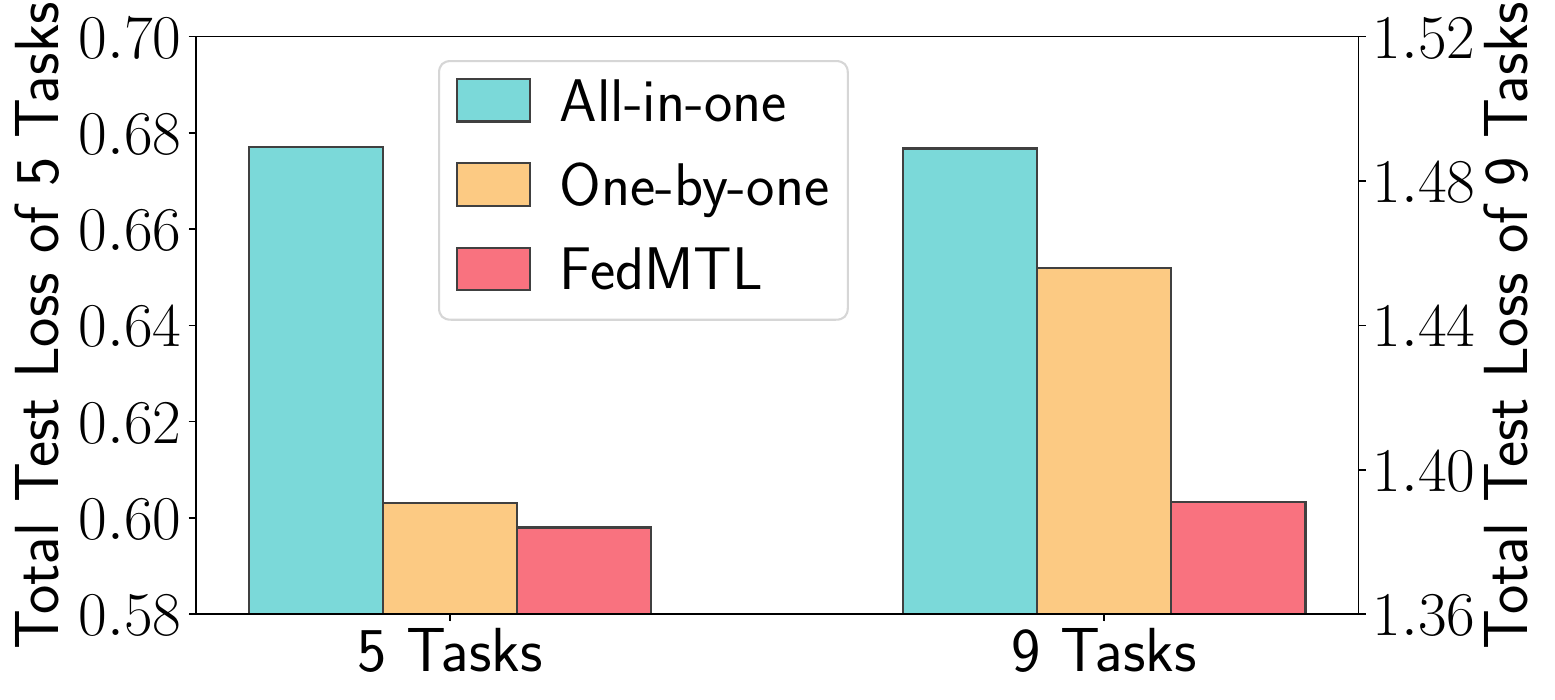}
     \vspace{-2ex}
	\caption{Federated multiple task learning with 5 tasks and 9 tasks.}
	\label{fig:fedmtl}
	\vspace{-3ex}
\end{figure}

\textbf{Federated Multiple-Task Learning:} We evaluate federated multiple-task learning via two sets of experiments with 5 and 9 different vision tasks, respectively. Two simple solutions to deal with multiple tasks are the \textit{all-in-one} method that uses a shared encoder for all tasks and the \textit{one-by-one} method that trains tasks sequentially. We benchmark them with FedMTL method \cite{zhuang2023mas}, which trains with all tasks combined first and then divides the tasks into groups for further training according to the task affinity. Figure~\ref{fig:fedmtl} shows that FedMTL not only can leverage the task-agnostic knowledge but also encourage collaboration between tasks with higher affinity to improve overall performance. We believe this sets a good foundation for future works to further improve upon in this nascent field.

\begin{figure}[t]
	\centering
	\includegraphics[width=0.81\columnwidth]{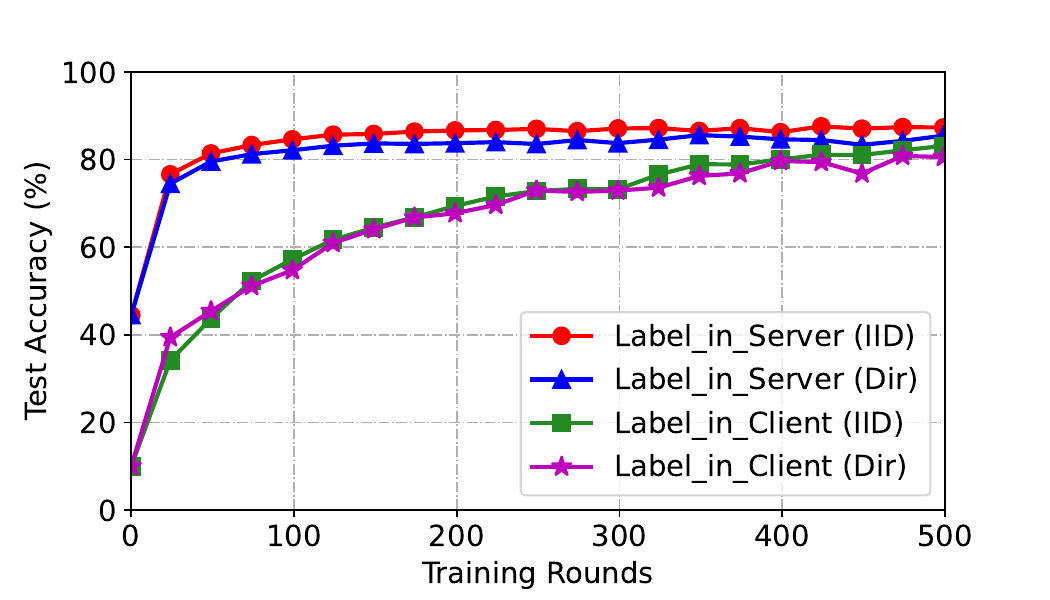}
 \vspace{-2ex}
	\caption{\small Benchmark results of semi-supervised FL.}
	\label{fig:cifar_semi}
		\vspace{-3ex}
\end{figure}

\subsection{Data Level Benchmark}
\textbf{Feature Shift:} We provide benchmark results for feature shift with multi-domain datasets, including Digits-5, Office-Caltech, and DomainNet. Each client exclusively contains samples from a specific domain. We follow the prior works \cite{li2020fedbn} and simulate the number of clients to be equal to the number of domains of each dataset. Table~\ref{tab:classification} shows that the FedAvg algorithm can already achieve a certain level of performance even in the presence of feature shifts, but there exist some performance gaps for different domains, as shown in Figure ~\ref{fig:multi-domain}.
The possible factors for this phenomenon could include the varied sample sizes, the underlying difficulty in discriminating objects in each domain, and the fairness of the learning algorithm. This prompts further research into developing fairer global models or more suitable personalized models.

\textbf{Semi-supervised Learning:} We benchmark both the label-in-server and the label-in-client scenarios for federated semi-supervised learning. The baseline methods are SemiFL~\cite{Diao22SemiFL} and FedAvg with FixMatch. We use the CIFAR-10 dataset for evaluation and the amount of labeled samples for each class is set as 400 for the label-in-server scenarios. For the label-in-client scenarios, we choose 10 labeled samples per class for each client. The remaining unlabeled samples among clients are partitioned either by random or Dirichlet-based allocation. Figure~\ref{fig:cifar_semi} shows that the label-in-server setting achieves better performance than the label-in-client setting as the global model fine-tuned by the server-side data is more unbiased and the pseudo-labels could be more reliable. 



\textbf{Self-supervised Learning:} We use the CIFAR-10/100 datasets to benchmark federated self-supervised learning with BOYL algorithm (FedBOYL) \cite{zhuang2021collaborative}. For the performance evaluation, we adopt the kNN evaluation on the outputs generated by the backbones. Table~\ref{tab:self-cifar} demonstrates that self-supervised representation learning is promising even in the FL settings.

\begin{table}[t]
		\small
	\centering
	\caption{Benchmark results of federated self-supervised learning based on BYOL method.}
		\renewcommand\arraystretch{0.9}
	\begin{tabular}{ l c c c c}
		\toprule[1pt]
		\textbf{Dataset} & \textbf{Method} & \textbf{\# client} &  IID  & Dir(0.5)   \\
		\midrule
		\multirow{2}{*}{CIFAR10} 
		& FedBOYL & 4 & 83.84 & 82.63
		\\
		& FedBOYL & 20 & 75.31 & 64.39
		\\
		\midrule
		\multirow{2}{*}{CIFAR100} 
		& FedBOYL & 4 & 48.65 & 47.19
		\\
		& FedBOYL & 20 & 39.48 & 38.69
		\\
		\bottomrule[1pt]
		\label{tab:self-cifar}
	\end{tabular}
		\vspace{-3ex}
\end{table}

\begin{figure}[t]
	\centering
	\subfigure[CIFAR10]{
		\includegraphics[width=0.47\columnwidth]{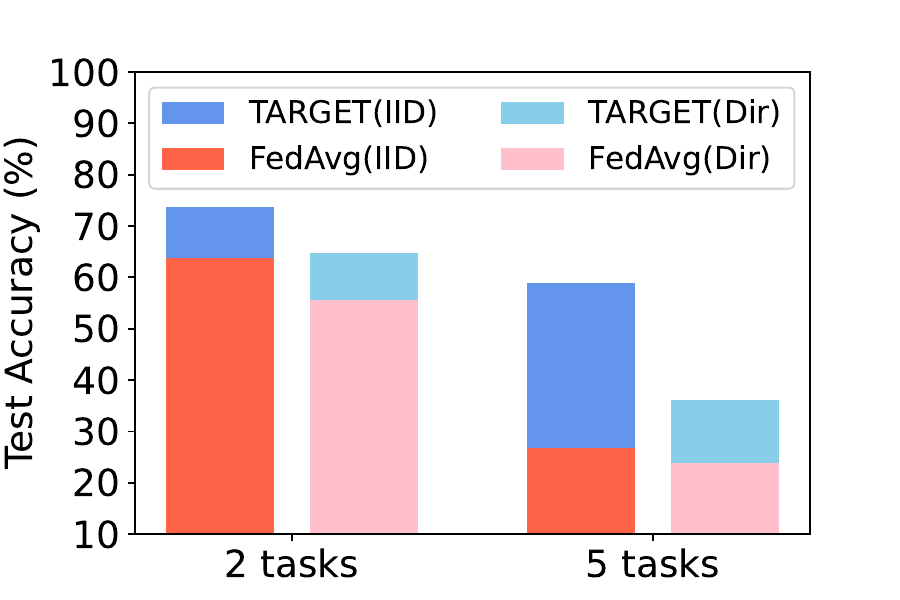}
	}
	\subfigure[CIFAR100]{
		\includegraphics[width=0.47\columnwidth]{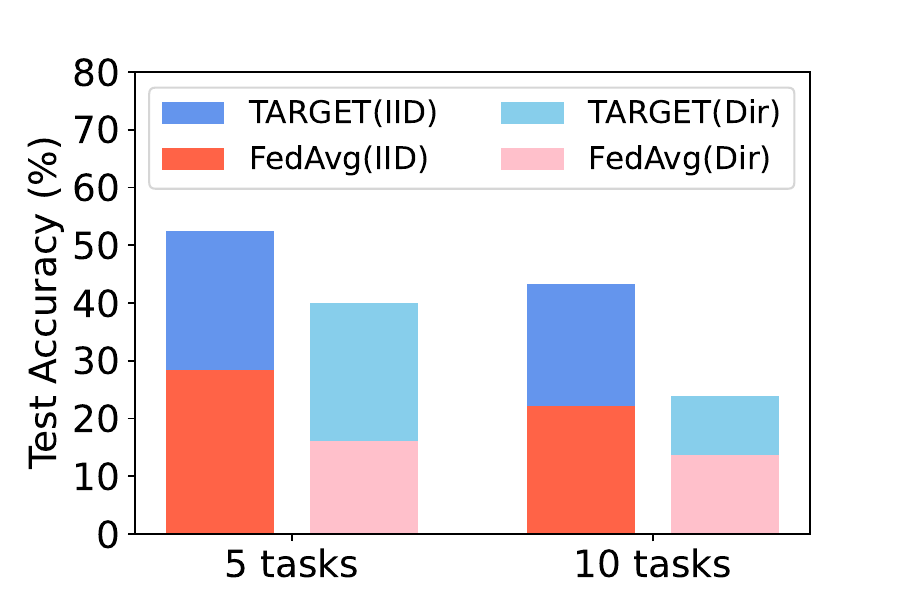}
	}
	    \vspace{-3ex}
	\caption{Results of class-incremental federated learning.}
	\label{fig:fccl}
		\vspace{-3ex}
\end{figure}

\textbf{Continual learning:} We benchmark federated continual learning with different numbers of incremental tasks on CIFAR-10/100 datasets, where each task contains a distinct set of classes. Both IID and non-IID label distribution are considered. We benchmark both FedAvg and TARGET \cite{zhang2023target}, which leverages both real data of the current task and synthetic data generated by investigating historical global model. The performance metric is the average accuracy of all tasks. Figure~\ref{fig:fccl} shows that synthetic data is effective in mitigating catastrophic forgetting issues.

\vspace{-0.5ex}

\begin{table}[t]
	\small
	\centering
	\caption{Benchmark results (\%) of test-time adaptation after FL.}
		\renewcommand\arraystretch{0.9}
	\begin{tabular}{ l c c c c c}
		\toprule[1pt]
		{\textbf{Dataset} }&  {\textbf{Shift Type}} & {\textbf{FedAvg}}&{\textbf{FedICON}}\\
		\midrule
		CIFAR10 (IID) & Original & 92.45 & 93.01
		\\
            CIFAR10 (IID) & Covariate & 59.41 & 64.07
		\\
		CIFAR10 (Dir) & Original & 89.22 & 92.38
		\\
		CIFAR10 (Dir) & Covariate & 53.43 & 76.62
		\\
		\midrule
		Digits-Five & Original & 94.16 & 95.21
            \\
            Digits-Five & Covariate & 89.63 & 95.27
		\\
            \midrule
		Office-Caltech & Original & 84.09 & 90.48
		\\
            Office-Caltech & Covariate & 65.02 & 76.27
            \\
		\bottomrule[1pt]
		\label{tab:test-time}
	\end{tabular}
		\vspace{-1ex}
\end{table}

\begin{table}[t]
         \vspace{-2ex}
	\small
	\centering
	\caption{Benchmark results (\%) of federated split learning.}
		\renewcommand\arraystretch{0.9}
	\begin{tabular}{ l c c c c}
		\toprule[1pt]
		\textbf{Dataset} & \textbf{Method} & \textbf{\# client} &  IID  & Dir(0.5)   \\
		\midrule
		\multirow{3}{*}{CIFAR10} 
		& MocoSFL & 100 & 79.23 & 79.45
		\\
		& MocoSFL & 1000 & 69.75 & 69.40
		\\
		\midrule
		\multirow{2}{*}{CIFAR100} 
		& MocoSFL & 100 &  43.29 & 43.30
		\\
		& MocoSFL & 1000 & 35.34 & 37.39
		\\
		\bottomrule[1pt]
		\label{tab:sfl-cifar}
	\end{tabular}
		\vspace{-1ex}
\end{table}

\begin{table}[t!]
	\vspace{-2ex}
	\small
	\centering
	\caption{Test accuracy (\%) of clustered and personalized FL.}
		\renewcommand\arraystretch{0.9}
	\begin{tabular}{ l c c c c}
		\toprule[1pt]
		\textbf{Datasets} & \textbf{FedAvg} &  \textbf{FedAvg-FT} &\textbf{IFCA}  &  \textbf{FedRep} \\
		\midrule
            CIFAR10 & 90.17 & 91.16 & 90.51 & 85.20
		\\
		Digits-5 & 94.74 & 94.97 & 95.60 & 94.14
		\\
		\bottomrule[1pt]
		\label{tab:multi-model}
	\end{tabular}
		\vspace{-1ex}
\end{table}

\begin{table}[t!]
	\vspace{-2ex}
	\small
	\centering
	\caption{Accuracy (\%) and percentage of trainable parameters (\% Params.) of fine-tuning ViT-B in FL.}
		\renewcommand\arraystretch{0.9}
	\begin{tabular}{ l c c c c}
		\toprule[1pt]
		\multirow{2}{*}{\textbf{Dataset} }& \multicolumn{2}{c}{\textbf{Fed-Linear}}&\multicolumn{2}{c}{\textbf{Fed-LoRA}}\\
		\cmidrule(lr){2-3}
            \cmidrule(lr){4-5}
		& Acc. & \% Params. & Acc. & \% Params. \\
		\midrule
		CIFAR10  (IID) & 95.36 & 0.0090 & 97.89 & 0.0519
		\\
            CIFAR10  (Dir) & 95.26 & 0.0090 & 97.89 & 0.0519
		\\
            CIFAR100 (IID) & 81.30 & 0.0895 & 90.57 & 0.1324
		\\
            CIFAR100 (Dir) & 80.50 & 0.0895 & 89.61 & 0.1324
		\\
            \midrule
		DomainNet & 88.45 & 0.0090 & 89.17 & 0.0519
		\\
		\bottomrule[1pt]
		\label{tab:lora}
	\end{tabular}
		\vspace{-4ex}
\end{table}

\textbf{Test-time Adaptation:} We consider the covariate shift during the 
test stage. Following \cite{tan2023taming}, we simulate the covariate shift by adding corruptions to the raw test images. We compare the vanilla FedAvg without any adaptation and FedICON \cite{tan2023taming} that applies the contrastive loss during feature representation learning and also conducts unsupervised model adaptation during the test stage. It can be seen from Table~\ref{tab:test-time} that in the multi-domain settings, applying contrastive loss for feature learning and local personalized classifier head for decision making can lead to performance improvement even without any data shift. FedICON can generally perform better than vanilla FedAvg in the presence of data distribution shifts. 


\subsection{Model Level Benchmark}

\textbf{Federated Split Learning:} To evaluate federated split learning, we employ the self-supervised MocoSFL \cite{Li23MocoSFL} on CIFAR-10/100 datasets. We provide benchmark results with two relatively large-scale setups, 100 and 1000 clients. Table~\ref{tab:sfl-cifar} shows that split training is capable of achieving high performance regardless of the data heterogeneity.

\textbf{Multiple Models with Clustered/Personalized FL:} We benchmark multiple models using the popular IFCA \cite{ghosh2020efficient} and FedRep \cite{collins2021fedrep}, as well as FedAvg. We simulate 30 clients on two datasets: Digits-5 dataset (with feature shifts) and CIFAR-10 dataset (with label shifts). The number of clusters is set to 5 by default. Table~\ref{tab:multi-model} demonstrates that multiple models can improve local performance than a single global model in the presence of data heterogeneity. In particular, IFCA may be more suitable for feature shift while local fine-tuning of the global model, e.g., FedAvg-FT, is promising in both settings.

\textbf{Federated Parameter-Efficient Fine-Tuning:} 
Our platform supports federated fine-tuning of foundation models. We compare a parameter-efficient fine-tuning (PEFT) method called LoRA  \cite{Hu22LoRA} (Fed-LoRA) with simply training a linear classifier head (Fed-Linear) on ViT-B/16 model \cite{Dosovitskiy21ViT} in FL. We set the rank of LoRA as r=1 and evaluate both fine-tuning methods with learning rates 5e-3 and 50 communication rounds. Table~\ref{tab:lora} shows that ViT is robust to data heterogeneity, which demonstrates the great potential of leveraging foundation models with PEFT techniques in FL.


\section{Conclusion}
In this paper, we present \name{}, a practical and vision-centric FL platform. It is highly customizable and automates benchmarking of a wide range of practical FL scenarios from multiple levels such as task, data, and model. For the task level, \name{} supports federated learning of 15 vision tasks and federated multiple-task learning. For the data level, \name{} covers heterogeneous feature and label distributions, continually changing data distribution, and learning with different degrees of annotation availability. For the model level, \name{} facilities federated split learning and federated learning of models with different architectures or parameters. The comprehensive benchmarking experiments under a wide range of FL scenarios validate the great potential of our proposed \name{}. We hope that \name{} sets a good foundation and will be useful to accelerate further advancements and landing of FL in various industry scenarios.

\section*{Impact Statement}
This work aims to largely advance the research and applications of federated learning in the area of computer vision. The developed platform is possible to help integrate cutting-edge research progress in FL while accelerating the landing of large-scale FL in the industry.

\bibliography{paper}

\begin{thebibliography}{88}
\providecommand{\natexlab}[1]{#1}
\providecommand{\url}[1]{\texttt{#1}}
\expandafter\ifx\csname urlstyle\endcsname\relax
  \providecommand{\doi}[1]{doi: #1}\else
  \providecommand{\doi}{doi: \begingroup \urlstyle{rm}\Url}\fi

\bibitem[Abadi et~al.(2016)Abadi, Agarwal, Barham, Brevdo, Chen, Citro, Corrado, Davis, Dean, Devin, et~al.]{abadi2016tensorflow-tensorboard}
Abadi, M., Agarwal, A., Barham, P., Brevdo, E., Chen, Z., Citro, C., Corrado, G.~S., Davis, A., Dean, J., Devin, M., et~al.
\newblock Tensorflow: Large-scale machine learning on heterogeneous distributed systems.
\newblock \emph{arXiv preprint arXiv:1603.04467}, 2016.

\bibitem[Alam et~al.(2023)Alam, Zhang, Feng, Shen, Cao, Zhao, Ko, Somasundaram, Narayanan, Avestimehr, et~al.]{alam2023fedaiot}
Alam, S., Zhang, T., Feng, T., Shen, H., Cao, Z., Zhao, D., Ko, J., Somasundaram, K., Narayanan, S.~S., Avestimehr, S., et~al.
\newblock Fedaiot: A federated learning benchmark for artificial intelligence of things.
\newblock \emph{arXiv preprint arXiv:2310.00109}, 2023.

\bibitem[Andriluka et~al.(2014)Andriluka, Pishchulin, Gehler, and Schiele]{andriluka14mpii}
Andriluka, M., Pishchulin, L., Gehler, P., and Schiele, B.
\newblock 2d human pose estimation: New benchmark and state of the art analysis.
\newblock In \emph{IEEE Conference on Computer Vision and Pattern Recognition (CVPR)}, June 2014.

\bibitem[Bao et~al.(2023)Bao, Wei, Wang, and He]{bao2023adaptive}
Bao, W., Wei, T., Wang, H., and He, J.
\newblock Adaptive test-time personalization for federated learning.
\newblock \emph{arXiv preprint arXiv:2310.18816}, 2023.

\bibitem[Beutel et~al.(2020)Beutel, Topal, Mathur, Qiu, Parcollet, and Lane]{beutel2020flower}
Beutel, D.~J., Topal, T., Mathur, A., Qiu, X., Parcollet, T., and Lane, N.~D.
\newblock Flower: A friendly federated learning research framework.
\newblock \emph{arXiv preprint arXiv:2007.14390}, 2020.

\bibitem[Bhuyan et~al.(2022)Bhuyan, Moharir, and Joshi]{bhuyan2022multi}
Bhuyan, N., Moharir, S., and Joshi, G.
\newblock Multi-model federated learning with provable guarantees.
\newblock In \emph{EAI International Conference on Performance Evaluation Methodologies and Tools}, pp.\  207--222. Springer, 2022.

\bibitem[Bonawitz et~al.(2019)Bonawitz, Eichner, Grieskamp, Huba, Ingerman, Ivanov, Kiddon, Kone\v{c}n\'{y}, Mazzocchi, McMahan, Van~Overveldt, Petrou, Ramage, and Roselander]{Bonawitz2019FL-sys-scale}
Bonawitz, K., Eichner, H., Grieskamp, W., Huba, D., Ingerman, A., Ivanov, V., Kiddon, C., Kone\v{c}n\'{y}, J., Mazzocchi, S., McMahan, B., Van~Overveldt, T., Petrou, D., Ramage, D., and Roselander, J.
\newblock Towards federated learning at scale: System design.
\newblock In Talwalkar, A., Smith, V., and Zaharia, M. (eds.), \emph{Proceedings of Machine Learning and Systems}. 2019.

\bibitem[Cai et~al.(2023)Cai, Wu, Wang, Lin, and Xu]{Cai23FedNLP}
Cai, D., Wu, Y., Wang, S., Lin, F.~X., and Xu, M.
\newblock Efficient federated learning for modern {NLP}.
\newblock pp.\  37:1--37:16. {ACM}, 2023.

\bibitem[Caldas et~al.(2018)Caldas, Duddu, Wu, Li, Kone{\v{c}}n{\`y}, McMahan, Smith, and Talwalkar]{leaf}
Caldas, S., Duddu, S. M.~K., Wu, P., Li, T., Kone{\v{c}}n{\`y}, J., McMahan, H.~B., Smith, V., and Talwalkar, A.
\newblock Leaf: A benchmark for federated settings.
\newblock \emph{arXiv preprint arXiv:1812.01097}, 2018.

\bibitem[Chen et~al.(2022)Chen, Gao, Kuang, Li, and Ding]{chen2022pflbench}
Chen, D., Gao, D., Kuang, W., Li, Y., and Ding, B.
\newblock p{FL}-bench: A comprehensive benchmark for personalized federated learning.
\newblock In \emph{Thirty-sixth Conference on Neural Information Processing Systems Datasets and Benchmarks Track}, 2022.
\newblock URL \url{https://openreview.net/forum?id=2ptbv_JjYKA}.

\bibitem[Chen et~al.(2023{\natexlab{a}})Chen, Gao, Xie, Pan, Li, Li, Ding, and Zhou]{chen2023FS-REAL}
Chen, D., Gao, D., Xie, Y., Pan, X., Li, Z., Li, Y., Ding, B., and Zhou, J.
\newblock {FS-REAL:} towards real-world cross-device federated learning.
\newblock In \emph{Proceedings of the 29th {ACM} {SIGKDD} Conference on Knowledge Discovery and Data Mining, {KDD} 2023, Long Beach, CA, USA, August 6-10, 2023}, pp.\  3829--3841. {ACM}, 2023{\natexlab{a}}.

\bibitem[Chen et~al.(2023{\natexlab{b}})Chen, Tan, Lu, Wu, and Hu]{chen2023openfed}
Chen, D., Tan, V.~J., Lu, Z., Wu, E., and Hu, J.
\newblock Openfed: A comprehensive and versatile open-source federated learning framework.
\newblock In \emph{Proceedings of the IEEE/CVF Conference on Computer Vision and Pattern Recognition}, pp.\  5017--5025, 2023{\natexlab{b}}.

\bibitem[Chen et~al.(2018)Chen, Zhu, Papandreou, Schroff, and Adam]{chen2018deeplab}
Chen, L.-C., Zhu, Y., Papandreou, G., Schroff, F., and Adam, H.
\newblock Encoder-decoder with atrous separable convolution for semantic image segmentation.
\newblock In \emph{Proceedings of the European conference on computer vision (ECCV)}, pp.\  801--818, 2018.

\bibitem[Collins et~al.(2021)Collins, Hassani, Mokhtari, and Shakkottai]{collins2021fedrep}
Collins, L., Hassani, H., Mokhtari, A., and Shakkottai, S.
\newblock Exploiting shared representations for personalized federated learning.
\newblock In \emph{International conference on machine learning}, pp.\  2089--2099. PMLR, 2021.

\bibitem[Diao et~al.(2021)Diao, Ding, and Tarokh]{Diao0T21HeteroFL}
Diao, E., Ding, J., and Tarokh, V.
\newblock Heterofl: Computation and communication efficient federated learning for heterogeneous clients.
\newblock In \emph{9th International Conference on Learning Representations, {ICLR} 2021, Virtual Event, Austria, May 3-7, 2021}, 2021.

\bibitem[Diao et~al.(2022)Diao, Ding, and Tarokh]{Diao22SemiFL}
Diao, E., Ding, J., and Tarokh, V.
\newblock Semifl: Semi-supervised federated learning for unlabeled clients with alternate training.
\newblock In \emph{NeurIPS}, 2022.

\bibitem[Dong et~al.(2022)Dong, Zhang, Li, and Kung]{dong2022spherefed}
Dong, X., Zhang, S.~Q., Li, A., and Kung, H.
\newblock Spherefed: Hyperspherical federated learning.
\newblock In \emph{European Conference on Computer Vision}, pp.\  165--184. Springer, 2022.

\bibitem[Dosovitskiy et~al.(2021)Dosovitskiy, Beyer, Kolesnikov, Weissenborn, Zhai, Unterthiner, Dehghani, Minderer, Heigold, Gelly, Uszkoreit, and Houlsby]{Dosovitskiy21ViT}
Dosovitskiy, A., Beyer, L., Kolesnikov, A., Weissenborn, D., Zhai, X., Unterthiner, T., Dehghani, M., Minderer, M., Heigold, G., Gelly, S., Uszkoreit, J., and Houlsby, N.
\newblock An image is worth 16x16 words: Transformers for image recognition at scale.
\newblock In \emph{9th International Conference on Learning Representations, {ICLR} 2021, Virtual Event, Austria, May 3-7, 2021}. OpenReview.net, 2021.

\bibitem[du~Terrail et~al.(2022)du~Terrail, Ayed, Cyffers, Grimberg, He, Loeb, Mangold, Marchand, Marfoq, Mushtaq, Muzellec, Philippenko, Silva, Tele{\'n}czuk, Albarqouni, Avestimehr, Bellet, Dieuleveut, Jaggi, Karimireddy, Lorenzi, Neglia, Tommasi, and Andreux]{terrail2022flamby}
du~Terrail, J.~O., Ayed, S.-S., Cyffers, E., Grimberg, F., He, C., Loeb, R., Mangold, P., Marchand, T., Marfoq, O., Mushtaq, E., Muzellec, B., Philippenko, C., Silva, S., Tele{\'n}czuk, M., Albarqouni, S., Avestimehr, S., Bellet, A., Dieuleveut, A., Jaggi, M., Karimireddy, S.~P., Lorenzi, M., Neglia, G., Tommasi, M., and Andreux, M.
\newblock {FL}amby: Datasets and benchmarks for cross-silo federated learning in realistic healthcare settings.
\newblock In \emph{Thirty-sixth Conference on Neural Information Processing Systems Datasets and Benchmarks Track}, 2022.
\newblock URL \url{https://openreview.net/forum?id=GgM5DiAb6A2}.

\bibitem[Everingham et~al.(2010)Everingham, Van~Gool, Williams, Winn, and Zisserman]{everingham2010pascal}
Everingham, M., Van~Gool, L., Williams, C.~K., Winn, J., and Zisserman, A.
\newblock The pascal visual object classes (voc) challenge.
\newblock \emph{International journal of computer vision}, 88:\penalty0 303--338, 2010.

\bibitem[Feng et~al.(2023)Feng, Bose, Zhang, Hebbar, Ramakrishna, Gupta, Zhang, Avestimehr, and Narayanan]{Feng23FedMultimodal}
Feng, T., Bose, D., Zhang, T., Hebbar, R., Ramakrishna, A., Gupta, R., Zhang, M., Avestimehr, S., and Narayanan, S.
\newblock Fedmultimodal: {A} benchmark for multimodal federated learning.
\newblock In \emph{Proceedings of the 29th {ACM} {SIGKDD} Conference on Knowledge Discovery and Data Mining, {KDD} 2023, Long Beach, CA, USA, August 6-10, 2023}, pp.\  4035--4045. {ACM}, 2023.

\bibitem[Garcia et~al.(2022)Garcia, Manoel, Diaz, Mireshghallah, Sim, and Dimitriadis]{garcia2022flute}
Garcia, M.~H., Manoel, A., Diaz, D.~M., Mireshghallah, F., Sim, R., and Dimitriadis, D.
\newblock Flute: A scalable, extensible framework for high-performance federated learning simulations.
\newblock \emph{arXiv preprint arXiv:2203.13789}, 2022.

\bibitem[Ghosh et~al.(2020)Ghosh, Chung, Yin, and Ramchandran]{ghosh2020efficient}
Ghosh, A., Chung, J., Yin, D., and Ramchandran, K.
\newblock An efficient framework for clustered federated learning.
\newblock \emph{Advances in Neural Information Processing Systems}, 33:\penalty0 19586--19597, 2020.

\bibitem[Grill et~al.(2020)Grill, Strub, Altch{\'e}, Tallec, Richemond, Buchatskaya, Doersch, Avila~Pires, Guo, Gheshlaghi~Azar, et~al.]{grill2020bootstrap}
Grill, J.-B., Strub, F., Altch{\'e}, F., Tallec, C., Richemond, P., Buchatskaya, E., Doersch, C., Avila~Pires, B., Guo, Z., Gheshlaghi~Azar, M., et~al.
\newblock Bootstrap your own latent-a new approach to self-supervised learning.
\newblock \emph{Advances in neural information processing systems}, 33:\penalty0 21271--21284, 2020.

\bibitem[He et~al.(2020)He, Li, So, Zhang, Wang, Wang, Vepakomma, Singh, Qiu, Shen, et~al.]{he2020fedml}
He, C., Li, S., So, J., Zhang, M., Wang, H., Wang, X., Vepakomma, P., Singh, A., Qiu, H., Shen, L., et~al.
\newblock Fedml: A research library and benchmark for federated machine learning.
\newblock \emph{arXiv preprint arXiv:2007.13518}, 2020.

\bibitem[He et~al.(2021)He, Shah, Tang, Fan, Sivashunmugam, Bhogaraju, Shimpi, Shen, Chu, Soltanolkotabi, and Avestimehr]{He21FedCV}
He, C., Shah, A.~D., Tang, Z., Fan, D., Sivashunmugam, A.~N., Bhogaraju, K., Shimpi, M., Shen, L., Chu, X., Soltanolkotabi, M., and Avestimehr, S.
\newblock Fedcv: {A} federated learning framework for diverse computer vision tasks.
\newblock \emph{CoRR}, abs/2111.11066, 2021.
\newblock URL \url{https://arxiv.org/abs/2111.11066}.

\bibitem[Hu et~al.(2022{\natexlab{a}})Hu, Shen, Wallis, Allen{-}Zhu, Li, Wang, Wang, and Chen]{Hu22LoRA}
Hu, E.~J., Shen, Y., Wallis, P., Allen{-}Zhu, Z., Li, Y., Wang, S., Wang, L., and Chen, W.
\newblock Lora: Low-rank adaptation of large language models.
\newblock In \emph{The Tenth International Conference on Learning Representations, {ICLR} 2022, Virtual Event, April 25-29, 2022}. OpenReview.net, 2022{\natexlab{a}}.

\bibitem[Hu et~al.(2020)Hu, Li, Liu, Li, Wu, and He]{hu2020oarf}
Hu, S., Li, Y., Liu, X., Li, Q., Wu, Z., and He, B.
\newblock The oarf benchmark suite: Characterization and implications for federated learning systems.
\newblock \emph{arXiv preprint arXiv:2006.07856}, 2020.

\bibitem[Hu et~al.(2022{\natexlab{b}})Hu, Li, Liu, Li, Wu, and He]{hu2022oarf}
Hu, S., Li, Y., Liu, X., Li, Q., Wu, Z., and He, B.
\newblock The oarf benchmark suite: Characterization and implications for federated learning systems.
\newblock \emph{ACM Transactions on Intelligent Systems and Technology (TIST)}, 13\penalty0 (4):\penalty0 1--32, 2022{\natexlab{b}}.

\bibitem[Huba et~al.(2022)Huba, Nguyen, Malik, Zhu, Rabbat, Yousefpour, Wu, Zhan, Ustinov, Srinivas, et~al.]{huba2022papaya}
Huba, D., Nguyen, J., Malik, K., Zhu, R., Rabbat, M., Yousefpour, A., Wu, C.-J., Zhan, H., Ustinov, P., Srinivas, H., et~al.
\newblock Papaya: Practical, private, and scalable federated learning.
\newblock \emph{Proceedings of Machine Learning and Systems}, 4:\penalty0 814--832, 2022.

\bibitem[Jeong et~al.(2021)Jeong, Yoon, Yang, and Hwang]{Jeong21FedMatch}
Jeong, W., Yoon, J., Yang, E., and Hwang, S.~J.
\newblock Federated semi-supervised learning with inter-client consistency {\&} disjoint learning.
\newblock In \emph{9th International Conference on Learning Representations, {ICLR}}. OpenReview.net, 2021.

\bibitem[Kairouz et~al.(2019)Kairouz, McMahan, Avent, Bellet, Bennis, Bhagoji, Bonawitz, Charles, Cormode, Cummings, et~al.]{kairouz2019fl-advances-open}
Kairouz, P., McMahan, H.~B., Avent, B., Bellet, A., Bennis, M., Bhagoji, A.~N., Bonawitz, K., Charles, Z., Cormode, G., Cummings, R., et~al.
\newblock Advances and open problems in federated learning.
\newblock \emph{arXiv preprint arXiv:1912.04977}, 2019.

\bibitem[Kim et~al.(2023)Kim, Lin, Lee, Lau, and Mugunthan]{kim2023navigating}
Kim, T., Lin, E., Lee, J., Lau, C., and Mugunthan, V.
\newblock Navigating data heterogeneity in federated learning: A semi-supervised approach for object detection.
\newblock In \emph{Thirty-seventh Conference on Neural Information Processing Systems}, 2023.
\newblock URL \url{https://openreview.net/forum?id=2D7ou48q0E}.

\bibitem[Lai et~al.(2022)Lai, Dai, Singapuram, Liu, Zhu, Madhyastha, and Chowdhury]{lai2022fedscale}
Lai, F., Dai, Y., Singapuram, S. S.~V., Liu, J., Zhu, X., Madhyastha, H.~V., and Chowdhury, M.
\newblock Fedscale: Benchmarking model and system performance of federated learning at scale.
\newblock In \emph{International Conference on Machine Learning, {ICML} 2022, 17-23 July 2022, Baltimore, Maryland, {USA}}, 2022.

\bibitem[Li \& Wang(2019)Li and Wang]{li2019fedmd}
Li, D. and Wang, J.
\newblock Fedmd: Heterogenous federated learning via model distillation.
\newblock \emph{arXiv preprint arXiv:1910.03581}, 2019.

\bibitem[Li et~al.(2023)Li, Lyu, Iso, Chakrabarti, and Spranger]{Li23MocoSFL}
Li, J., Lyu, L., Iso, D., Chakrabarti, C., and Spranger, M.
\newblock Mocosfl: enabling cross-client collaborative self-supervised learning.
\newblock In \emph{The Eleventh International Conference on Learning Representations, {ICLR}}. OpenReview.net, 2023.

\bibitem[Li et~al.(2022)Li, Diao, Chen, and He]{li2022federated}
Li, Q., Diao, Y., Chen, Q., and He, B.
\newblock Federated learning on non-iid data silos: An experimental study.
\newblock In \emph{2022 IEEE 38th International Conference on Data Engineering (ICDE)}, pp.\  965--978. IEEE, 2022.

\bibitem[Li et~al.(2020{\natexlab{a}})Li, Sahu, Zaheer, Sanjabi, Talwalkar, and Smith]{fedprox}
Li, T., Sahu, A.~K., Zaheer, M., Sanjabi, M., Talwalkar, A., and Smith, V.
\newblock Federated optimization in heterogeneous networks.
\newblock In \emph{Proceedings of Machine Learning and Systems 2020}, pp.\  429--450. 2020{\natexlab{a}}.

\bibitem[Li et~al.(2020{\natexlab{b}})Li, JIANG, Zhang, Kamp, and Dou]{li2020fedbn}
Li, X., JIANG, M., Zhang, X., Kamp, M., and Dou, Q.
\newblock Fedbn: Federated learning on non-iid features via local batch normalization.
\newblock In \emph{International Conference on Learning Representations}, 2020{\natexlab{b}}.

\bibitem[Liang et~al.(2022)Liang, Lin, Fu, Zhu, and Li]{Liang22RSCFed}
Liang, X., Lin, Y., Fu, H., Zhu, L., and Li, X.
\newblock Rscfed: Random sampling consensus federated semi-supervised learning.
\newblock In \emph{{IEEE/CVF} Conference on Computer Vision and Pattern Recognition, {CVPR}}, 2022.

\bibitem[Lin et~al.(2022)Lin, He, Ze, Wang, Hua, Dupuy, Gupta, Soltanolkotabi, Ren, and Avestimehr]{Lin22FedNLP}
Lin, B.~Y., He, C., Ze, Z., Wang, H., Hua, Y., Dupuy, C., Gupta, R., Soltanolkotabi, M., Ren, X., and Avestimehr, S.
\newblock Fednlp: Benchmarking federated learning methods for natural language processing tasks.
\newblock In \emph{Findings of the Association for Computational Linguistics: {NAACL} 2022, Seattle, WA, United States, July 10-15, 2022}, 2022.

\bibitem[Lin et~al.(2020)Lin, Kong, Stich, and Jaggi]{lin2020ensemble}
Lin, T., Kong, L., Stich, S.~U., and Jaggi, M.
\newblock Ensemble distillation for robust model fusion in federated learning.
\newblock \emph{Advances in Neural Information Processing Systems}, 33:\penalty0 2351--2363, 2020.

\bibitem[Liu et~al.(2022)Liu, Wang, Chien, and Lai]{Liu22FedFR}
Liu, C., Wang, C., Chien, S., and Lai, S.
\newblock Fedfr: Joint optimization federated framework for generic and personalized face recognition.
\newblock In \emph{Thirty-Sixth {AAAI} Conference on Artificial Intelligence, {AAAI}}, 2022.

\bibitem[Liu et~al.(2020)Liu, Huang, Luo, Huang, Liu, Chen, Feng, Chen, Yu, and Yang]{liu2020fedvision}
Liu, Y., Huang, A., Luo, Y., Huang, H., Liu, Y., Chen, Y., Feng, L., Chen, T., Yu, H., and Yang, Q.
\newblock Fedvision: An online visual object detection platform powered by federated learning.
\newblock In \emph{Proceedings of the AAAI conference on artificial intelligence}, volume~34, pp.\  13172--13179, 2020.

\bibitem[Luo et~al.(2022)Luo, Xiao, and Song]{Luo22pFLRec}
Luo, S., Xiao, Y., and Song, L.
\newblock Personalized federated recommendation via joint representation learning, user clustering, and model adaptation.
\newblock In \emph{Proceedings of the 31st {ACM} International Conference on Information {\&} Knowledge Management, Atlanta, GA, USA, October 17-21, 2022}, pp.\  4289--4293. {ACM}, 2022.

\bibitem[McMahan et~al.(2017)McMahan, Moore, Ramage, Hampson, and y~Arcas]{fedavg}
McMahan, B., Moore, E., Ramage, D., Hampson, S., and y~Arcas, B.~A.
\newblock Communication-efficient learning of deep networks from decentralized data.
\newblock In \emph{Artificial Intelligence and Statistics}, pp.\  1273--1282. PMLR, 2017.

\bibitem[Meng et~al.(2022)Meng, Zhou, Ren, Feng, Liu, and Lin]{Meng22Face}
Meng, Q., Zhou, F., Ren, H., Feng, T., Liu, G., and Lin, Y.
\newblock Improving federated learning face recognition via privacy-agnostic clusters.
\newblock In \emph{The Tenth International Conference on Learning Representations, {ICLR}}. OpenReview.net, 2022.

\bibitem[Miao et~al.(2023)Miao, Yang, Fan, and Yang]{miao2023fedseg}
Miao, J., Yang, Z., Fan, L., and Yang, Y.
\newblock Fedseg: Class-heterogeneous federated learning for semantic segmentation.
\newblock In \emph{Proceedings of the IEEE/CVF Conference on Computer Vision and Pattern Recognition}, pp.\  8042--8052, 2023.

\bibitem[Nguyen et~al.(2022)Nguyen, Pham, Pathirana, Ding, Seneviratne, Lin, Dobre, and Hwang]{nguyen2022fedhealth}
Nguyen, D.~C., Pham, Q.-V., Pathirana, P.~N., Ding, M., Seneviratne, A., Lin, Z., Dobre, O., and Hwang, W.-J.
\newblock Federated learning for smart healthcare: A survey.
\newblock \emph{ACM Computing Surveys (CSUR)}, 55\penalty0 (3):\penalty0 1--37, 2022.

\bibitem[Niu \& Deng(2022)Niu and Deng]{NiuD22fedface}
Niu, Y. and Deng, W.
\newblock Federated learning for face recognition with gradient correction.
\newblock In \emph{Thirty-Sixth {AAAI} Conference on Artificial Intelligence, {AAAI}}, 2022.

\bibitem[PaddlePaddle(2020)]{paddlefl-github}
PaddlePaddle.
\newblock Paddlefl vanilla implementation, 2020.
\newblock URL \url{https://github.com/PaddlePaddle/PaddleFL/tree/master/python/paddle_fl/paddle_fl/examples/femnist_demo}.

\bibitem[Qin et~al.(2023)Qin, Deng, Zhao, and Yan]{qin2023fedapen}
Qin, Z., Deng, S., Zhao, M., and Yan, X.
\newblock Fedapen: Personalized cross-silo federated learning with adaptability to statistical heterogeneity.
\newblock In \emph{Proceedings of the 29th ACM SIGKDD Conference on Knowledge Discovery and Data Mining}, pp.\  1954--1964, 2023.

\bibitem[Reddi et~al.(2020)Reddi, Charles, Zaheer, Garrett, Rush, Kone{\v{c}}n{\`y}, Kumar, and McMahan]{reddi2020fedyogi}
Reddi, S., Charles, Z., Zaheer, M., Garrett, Z., Rush, K., Kone{\v{c}}n{\`y}, J., Kumar, S., and McMahan, H.~B.
\newblock Adaptive federated optimization.
\newblock \emph{arXiv preprint arXiv:2003.00295}, 2020.

\bibitem[Redmon et~al.(2016)Redmon, Divvala, Girshick, and Farhadi]{redmon2016yolo}
Redmon, J., Divvala, S., Girshick, R., and Farhadi, A.
\newblock You only look once: Unified, real-time object detection.
\newblock In \emph{Proceedings of the IEEE conference on computer vision and pattern recognition}, pp.\  779--788, 2016.

\bibitem[Ren et~al.(2015)Ren, He, Girshick, and Sun]{ren2015faster}
Ren, S., He, K., Girshick, R., and Sun, J.
\newblock Faster r-cnn: Towards real-time object detection with region proposal networks.
\newblock \emph{Advances in neural information processing systems}, 28, 2015.

\bibitem[Ryffel et~al.(2018)Ryffel, Trask, Dahl, Wagner, Mancuso, Rueckert, and Passerat-Palmbach]{pysyft}
Ryffel, T., Trask, A., Dahl, M., Wagner, B., Mancuso, J., Rueckert, D., and Passerat-Palmbach, J.
\newblock A generic framework for privacy preserving deep learning.
\newblock \emph{arXiv preprint arXiv:1811.04017}, 2018.

\bibitem[Tan et~al.(2023)Tan, Chen, Zhuang, Dong, Lyu, and Long]{tan2023taming}
Tan, Y., Chen, C., Zhuang, W., Dong, X., Lyu, L., and Long, G.
\newblock Taming heterogeneity to deal with test-time shift in federated learning.
\newblock In \emph{International Workshop on Federated Learning for Distributed Data Mining}, 2023.

\bibitem[Tan et~al.(2024)Tan, Chen, Zhuang, Dong, Lyu, and Long]{tan2024heterogeneity}
Tan, Y., Chen, C., Zhuang, W., Dong, X., Lyu, L., and Long, G.
\newblock Is heterogeneity notorious? taming heterogeneity to handle test-time shift in federated learning.
\newblock \emph{Advances in Neural Information Processing Systems}, 36, 2024.

\bibitem[Tensorflow.org(2019)]{tff}
Tensorflow.org.
\newblock Tensorflow federated, 2019.
\newblock URL \url{https://github.com/tensorflow/federated}.

\bibitem[Thapa et~al.(2022)Thapa, Arachchige, Camtepe, and Sun]{thapa2022splitfed}
Thapa, C., Arachchige, P. C.~M., Camtepe, S., and Sun, L.
\newblock Splitfed: When federated learning meets split learning.
\newblock In \emph{Proceedings of the AAAI Conference on Artificial Intelligence}, volume~36, pp.\  8485--8493, 2022.

\bibitem[Toshev \& Szegedy(2014)Toshev and Szegedy]{toshev2014deeppose}
Toshev, A. and Szegedy, C.
\newblock Deeppose: Human pose estimation via deep neural networks.
\newblock In \emph{Proceedings of the IEEE conference on computer vision and pattern recognition}, pp.\  1653--1660, 2014.

\bibitem[Wang et~al.(2021)Wang, Deng, Meng, Wang, Li, Lin, Han, Miao, Rajasekaran, and Ding]{Wang21FLNLP}
Wang, C., Deng, J., Meng, X., Wang, Y., Li, J., Lin, S., Han, S., Miao, F., Rajasekaran, S., and Ding, C.
\newblock A secure and efficient federated learning framework for {NLP}.
\newblock In \emph{Proceedings of the 2021 Conference on Empirical Methods in Natural Language Processing, {EMNLP} 2021, Virtual Event / Punta Cana, Dominican Republic, 7-11 November, 2021}, pp.\  7676--7682. Association for Computational Linguistics, 2021.

\bibitem[Wang et~al.(2023{\natexlab{a}})Wang, Chen, Chowdhury, Kannan, and Liang]{wang2023flint}
Wang, E., Chen, B., Chowdhury, M., Kannan, A., and Liang, F.
\newblock Flint: A platform for federated learning integration.
\newblock \emph{Proceedings of Machine Learning and Systems}, 5, 2023{\natexlab{a}}.

\bibitem[Wang et~al.(2023{\natexlab{b}})Wang, Fan, Peng, Li, Yang, Feng, Yang, Liu, and Wang]{wang2023flgo}
Wang, Z., Fan, X., Peng, Z., Li, X., Yang, Z., Feng, M., Yang, Z., Liu, X., and Wang, C.
\newblock Flgo: A fully customizable federated learning platform.
\newblock \emph{arXiv preprint arXiv:2306.12079}, 2023{\natexlab{b}}.

\bibitem[WeBank(2019)]{fate}
WeBank.
\newblock Federated ai technology enabler (fate), 2019.
\newblock URL \url{https://github.com/FederatedAI/FATE}.

\bibitem[Woisetschl{\"{a}}ger et~al.(2024)Woisetschl{\"{a}}ger, Isenko, Wang, Mayer, and Jacobsen]{Herbert24SurveyFLFM}
Woisetschl{\"{a}}ger, H., Isenko, A., Wang, S., Mayer, R., and Jacobsen, H.
\newblock A survey on efficient federated learning methods for foundation model training.
\newblock \emph{CoRR}, abs/2401.04472, 2024.

\bibitem[Xie et~al.(2023)Xie, Wang, Gao, Chen, Yao, Kuang, Li, Ding, and Zhou]{Xie23FederatedScope}
Xie, Y., Wang, Z., Gao, D., Chen, D., Yao, L., Kuang, W., Li, Y., Ding, B., and Zhou, J.
\newblock Federatedscope: {A} flexible federated learning platform for heterogeneity.
\newblock \emph{Proc. {VLDB} Endow.}, 16\penalty0 (5):\penalty0 1059--1072, 2023.

\bibitem[Yang et~al.(2023)Yang, Xu, Ding, and Liu]{yang2023fedhap}
Yang, M., Xu, J., Ding, W., and Liu, Y.
\newblock Fedhap: Federated hashing with global prototypes for cross-silo retrieval.
\newblock \emph{IEEE Transactions on Parallel and Distributed Systems}, 2023.

\bibitem[Yu et~al.(2020)Yu, Chen, Wang, Xian, Chen, Liu, Madhavan, and Darrell]{yu2020bdd100k}
Yu, F., Chen, H., Wang, X., Xian, W., Chen, Y., Liu, F., Madhavan, V., and Darrell, T.
\newblock Bdd100k: A diverse driving dataset for heterogeneous multitask learning.
\newblock In \emph{Proceedings of the IEEE/CVF conference on computer vision and pattern recognition}, pp.\  2636--2645, 2020.

\bibitem[Zamir et~al.(2018)Zamir, Sax, Shen, Guibas, Malik, and Savarese]{zamir2018taskonomy}
Zamir, A.~R., Sax, A., Shen, W., Guibas, L.~J., Malik, J., and Savarese, S.
\newblock Taskonomy: Disentangling task transfer learning.
\newblock In \emph{Proceedings of the IEEE conference on computer vision and pattern recognition}, pp.\  3712--3722, 2018.

\bibitem[Zeng et~al.(2023)Zeng, Liang, Hu, Wang, and Xu]{Zeng23FedLab}
Zeng, D., Liang, S., Hu, X., Wang, H., and Xu, Z.
\newblock Fedlab: {A} flexible federated learning framework.
\newblock \emph{J. Mach. Learn. Res.}, 24:\penalty0 100:1--100:7, 2023.

\bibitem[Zhang et~al.(2023{\natexlab{a}})Zhang, Long, Zhou, Yan, Zhang, Zhang, and Yang]{Zhang23dupflrec}
Zhang, C., Long, G., Zhou, T., Yan, P., Zhang, Z., Zhang, C., and Yang, B.
\newblock Dual personalization on federated recommendation.
\newblock In \emph{Proceedings of the Thirty-Second International Joint Conference on Artificial Intelligence, {IJCAI} 2023, 19th-25th August 2023, Macao, SAR, China}, 2023{\natexlab{a}}.

\bibitem[Zhang et~al.(2023{\natexlab{b}})Zhang, Chen, Zhuang, and Lyu]{zhang2023target}
Zhang, J., Chen, C., Zhuang, W., and Lyu, L.
\newblock Target: Federated class-continual learning via exemplar-free distillation.
\newblock In \emph{Proceedings of the IEEE/CVF International Conference on Computer Vision}, pp.\  4782--4793, 2023{\natexlab{b}}.

\bibitem[Zhang et~al.(2022)Zhang, Wu, Zhou, Zhou, Yang, and Jin]{Zhang22Felicitas}
Zhang, Q., Wu, T., Zhou, P., Zhou, S., Yang, Y., and Jin, X.
\newblock Felicitas: Federated learning in distributed cross device collaborative frameworks.
\newblock In Zhang, A. and Rangwala, H. (eds.), \emph{{KDD} '22: The 28th {ACM} {SIGKDD} Conference on Knowledge Discovery and Data Mining, Washington, DC, USA, August 14 - 18, 2022}, pp.\  4502--4509. {ACM}, 2022.

\bibitem[Zhang et~al.(2023{\natexlab{c}})Zhang, Feng, Alam, Lee, Zhang, Narayanan, and Avestimehr]{zhang2023fedaudio}
Zhang, T., Feng, T., Alam, S., Lee, S., Zhang, M., Narayanan, S.~S., and Avestimehr, S.
\newblock Fedaudio: A federated learning benchmark for audio tasks.
\newblock In \emph{ICASSP 2023-2023 IEEE International Conference on Acoustics, Speech and Signal Processing (ICASSP)}, pp.\  1--5. IEEE, 2023{\natexlab{c}}.

\bibitem[Zhang et~al.(2021)Zhang, Yang, Yao, Yan, Gonzalez, Ramchandran, and Mahoney]{zhang2021improving}
Zhang, Z., Yang, Y., Yao, Z., Yan, Y., Gonzalez, J.~E., Ramchandran, K., and Mahoney, M.~W.
\newblock Improving semi-supervised federated learning by reducing the gradient diversity of models.
\newblock In \emph{2021 IEEE International Conference on Big Data (Big Data)}, pp.\  1214--1225. IEEE, 2021.

\bibitem[Zhang et~al.(2023{\natexlab{d}})Zhang, Yang, Dai, Wang, Yu, Qu, and Xu]{Zhang23FedPETuning}
Zhang, Z., Yang, Y., Dai, Y., Wang, Q., Yu, Y., Qu, L., and Xu, Z.
\newblock Fedpetuning: When federated learning meets the parameter-efficient tuning methods of pre-trained language models.
\newblock In Rogers, A., Boyd{-}Graber, J.~L., and Okazaki, N. (eds.), \emph{Findings of the Association for Computational Linguistics: {ACL} 2023, Toronto, Canada, July 9-14, 2023}, pp.\  9963--9977. Association for Computational Linguistics, 2023{\natexlab{d}}.

\bibitem[Zhuang \& Lyu(2023)Zhuang and Lyu]{zhuang2023normalization}
Zhuang, W. and Lyu, L.
\newblock Is normalization indispensable for multi-domain federated learning?
\newblock \emph{arXiv preprint arXiv:2306.05879}, 2023.

\bibitem[Zhuang \& Lyu(2024)Zhuang and Lyu]{zhuang2024fedwon}
Zhuang, W. and Lyu, L.
\newblock Fedwon: Triumphing multi-domain federated learning without normalization.
\newblock In \emph{The Twelfth International Conference on Learning Representations, {ICLR}}, 2024.
\newblock URL \url{https://openreview.net/forum?id=hAYHmV1gM8}.

\bibitem[Zhuang et~al.(2020)Zhuang, Wen, Zhang, Gan, Yin, Zhou, Zhang, and Yi]{zhuang2020fedreid}
Zhuang, W., Wen, Y., Zhang, X., Gan, X., Yin, D., Zhou, D., Zhang, S., and Yi, S.
\newblock Performance optimization of federated person re-identification via benchmark analysis.
\newblock In \emph{Proceedings of the 28th ACM International Conference on Multimedia}, pp.\  955--963, 2020.

\bibitem[Zhuang et~al.(2021{\natexlab{a}})Zhuang, Gan, Wen, Zhang, and Yi]{zhuang2021collaborative}
Zhuang, W., Gan, X., Wen, Y., Zhang, S., and Yi, S.
\newblock Collaborative unsupervised visual representation learning from decentralized data.
\newblock In \emph{Proceedings of the IEEE/CVF international conference on computer vision}, pp.\  4912--4921, 2021{\natexlab{a}}.

\bibitem[Zhuang et~al.(2021{\natexlab{b}})Zhuang, Wen, and Zhang]{zhuang2021joint}
Zhuang, W., Wen, Y., and Zhang, S.
\newblock Joint optimization in edge-cloud continuum for federated unsupervised person re-identification.
\newblock In \emph{Proceedings of the 29th ACM International Conference on Multimedia}, pp.\  433--441, 2021{\natexlab{b}}.

\bibitem[Zhuang et~al.(2022{\natexlab{a}})Zhuang, Gan, Wen, and Zhang]{Zhuang22EasyFL}
Zhuang, W., Gan, X., Wen, Y., and Zhang, S.
\newblock Easyfl: {A} low-code federated learning platform for dummies.
\newblock \emph{{IEEE} Internet Things J.}, 9\penalty0 (15):\penalty0 13740--13754, 2022{\natexlab{a}}.

\bibitem[Zhuang et~al.(2022{\natexlab{b}})Zhuang, Gan, Zhang, Wen, Zhang, and Yi]{zhuang2022fedfr}
Zhuang, W., Gan, X., Zhang, X., Wen, Y., Zhang, S., and Yi, S.
\newblock Federated unsupervised domain adaptation for face recognition.
\newblock In \emph{2022 IEEE International Conference on Multimedia and Expo (ICME)}, pp.\  1--6. IEEE, 2022{\natexlab{b}}.

\bibitem[Zhuang et~al.(2022{\natexlab{c}})Zhuang, Wen, and Zhang]{Zhuang22FedEMA}
Zhuang, W., Wen, Y., and Zhang, S.
\newblock Divergence-aware federated self-supervised learning.
\newblock In \emph{The Tenth International Conference on Learning Representations, {ICLR}}. OpenReview.net, 2022{\natexlab{c}}.

\bibitem[Zhuang et~al.(2023{\natexlab{a}})Zhuang, Chen, and Lyu]{zhuang23FMFL}
Zhuang, W., Chen, C., and Lyu, L.
\newblock When foundation model meets federated learning: Motivations, challenges, and future directions.
\newblock \emph{CoRR}, abs/2306.15546, 2023{\natexlab{a}}.
\newblock URL \url{https://doi.org/10.48550/arXiv.2306.15546}.

\bibitem[Zhuang et~al.(2023{\natexlab{b}})Zhuang, Gan, Wen, and Zhang]{zhuang2023optimizing}
Zhuang, W., Gan, X., Wen, Y., and Zhang, S.
\newblock Optimizing performance of federated person re-identification: Benchmarking and analysis.
\newblock \emph{ACM Transactions on Multimedia Computing, Communications and Applications}, 19\penalty0 (1s):\penalty0 1--18, 2023{\natexlab{b}}.

\bibitem[Zhuang et~al.(2023{\natexlab{c}})Zhuang, Wen, Lyu, and Zhang]{zhuang2023mas}
Zhuang, W., Wen, Y., Lyu, L., and Zhang, S.
\newblock Mas: Towards resource-efficient federated multiple-task learning.
\newblock In \emph{Proceedings of the IEEE/CVF International Conference on Computer Vision}, pp.\  23414--23424, 2023{\natexlab{c}}.

\end{thebibliography}
\bibliographystyle{icml2024}

\newpage
\appendix
\onecolumn

\section{Introduction of Representative Datasets}

\textbf{Digits-Five} \cite{li2020fedbn} is an integrated dataset consisting of five digits recognition sub-sets, including MNIST, MNIST-M, Synthetic Digits, SVHN, and USPS, each of which has a special style of digits from 0 to 9.

\textbf{Office-Caltech-10} \cite{li2020fedbn} consists of 10-category data samples from four different sources, including the Amazon merchant website, the Caltech-101 dataset, a high-resolution DSLR camera, and a webcam. Due to distinct feature distributions, each data source can represent a specific domain. There is no pre-split of training and test parts for office-caltech, so we randomly split the samples in each domain, of which 70\% are selected as training samples and the remainder as test samples.

\textbf{DomainNet} \cite{li2020fedbn} is a popular multi-domain dataset containing images from six distinct domains, including Clipart, Infograph, Painting, Quickdraw, Real, and Sketch. It has 345 object categories and we choose the top ten most common classes for our experiments as in \cite{li2020fedbn}.

\textbf{BDD100K} \cite{yu2020bdd100k} is a large-scale, diverse driving video database designed for research on autonomous vehicles and computer vision tasks. In this platform, we mainly make use the images for object detection. It covers a wide range of driving scenarios including weather conditions and urban environments. A variety of object categories, including pedestrians, cyclists, vehicles, etc, are annotated for evaluation.

\textbf{Pascal VOC} \cite{everingham2010pascal} (Visual Object Classes) dataset is a widely used benchmark in computer vision for object detection and image segmentation tasks. It covers common categories such as people, animals, vehicles, and household items.

\textbf{MPII} \cite{andriluka14mpii}. MPII Human Pose dataset is a popular benchmark for articulated human pose estimation. The dataset includes around 25K images containing over 40K people with annotated body joints. The images were systematically collected using an established taxonomy of every day human activities.


\begin{table*}[ht]
	\small
	\centering
	\caption{Currently tasks, datasets and scenarios supported by our platform. `Sup' means `Supervised'. \textbf{For each task, the dataset list can be easily extended by integrating new datasets}.}
		\renewcommand\arraystretch{1.3}
	\begin{tabular}{ l c c c}
		\toprule[1pt]
		\textbf{Task} & \textbf{Dataset} & \textbf{Split} & \textbf{Annotation}\\
		\midrule
		  \multirow{2}{*}{Image Classification} & CIFAR10/100, FEMNIST, Landmarks & IID, Dir, \# of classes & Sup, Semi-Sup, Self-sup
		\\
            & Digits-Five, Office-Caltech, DomainNet & Multi-Domain & Sup
            \\
            \midrule
		Object Detection & BDD100K & IID, Dir, H-Dir & Sup, Semi-Sup
		\\
           \midrule
            Semantic Segmentation & Pasacal VOC & IID, Dir, \# of classes & Sup
		\\
           \midrule
		Pose Estimation & MPII  & IID & Sup
		\\
        \midrule
            \multirow{2}{*}{Person Re-identification} & MSMT17, Market-1501, CUHK01, CUHK03-NP  & \multirow{2}{*}{Multi-Domain} & \multirow{2}{*}{Sup}
    		\\
            & PRID2011, VIPeR, 3DPeS, iLIDS-VID \\
            \midrule
		Face Recognition & BUPT-Balancedface, MS-Celeb-1M, WebFace, RFW  & Multi-Domain & Sup
		\\
        \midrule
		  3D-Point Cloud & ModelNet40 & IID & Sup
		\\
         \midrule
		  Multiple Tasks $^*$ &  Taskonomy & IID, Non-IID & Sup
		\\
		\bottomrule[1pt]
		\label{tab:overview}
 \end{tabular}

 \vspace{-2.9ex}
 
 \footnotesize $^*$ Multiple tasks training includes semantic segmentation, depth estimation, surface normal, keypoint, edge texture, \\ edge occlusion, reshaping, principle curvature, auto-encoder
\end{table*}

\section{Details of Experimental Settings} \label{app:exp_setup}

\textbf{Data Split.} In our benchmark experiments, we use the default training-test data split supported in the platform as described in Table~\ref{tab:benchmarks}. Basically, we make use of all the available training data, except the Digits5 and DomainNet, where we make some sampling to select only an identical amount of training data across different domains as in \cite{li2020fedbn}. 

\textbf{Model Selection.} We provide some predefined model architectures for each set of experiments. For example, simple CNN for Digits-5, ResNet-18 for CIFAR-10/100, AlexNet for Office-Caltech and DomainNet, YOLO-V5 series for Object Detection in BDD100K, etc. For most experiments, we train the global model from scratch as in conventional FL studies. For semantic segmentation and pose estimation, we utilize the pre-trained backbones.

\textbf{Training Hype-parameters.} For local training, SGD is selected as the default local optimizer with mini-batch size 32, learning rate 0.01, momentum 0.9, weight decay 0.0005 and local epoch $E$ = 5 unless otherwise mentioned. The number of communication rounds is set to 100 for three multi-domain datasets and 200 for other datasets. In particular, we extend the communication rounds in semi-supervised training to 500 as it requires more epochs for convergence. For parameter-efficient fine-tuning of foundation models, we set the local epoch $E$ = 1 as ViT is more computational intensive and the number of communication rounds as 50 as it can converge much faster than training from scratch of other CNN-based models. For all experiments, we report the best test metrics achieved for performance evaluation.

\textbf{System Scale.} For different tasks, the numbers of clients and participating rate are adjusted according to the total training data size. For Digits5, Office-Caltech and DomainNet, by default we allocate data from each specific domain to one client for simplicity. For CIFAR-10/100, we consider 100 clients with 10\% participation rate in supervised/semi-supervised learning and 4/20 clients with full participation for self-supervised learning as contrastive learning methods usually need large batch size. All experiments are run on a AWS cloud server equipped with four V100 GPUs.

\begin{table}[ht]
	\centering
	\caption{Dataset names, training and test sample sizes and used models for benchmark evaluation.}
 
		\renewcommand\arraystretch{1.1}
	\begin{tabular}{ l c c c c}
		\toprule[1pt]
		{Datasets} & Training Size  & Test Size & Models & Metric\\
		\midrule
		CIFAR10/100 & 50,000  & 10,000 & ResNet-18 & Accuracy
		\\
		Digits-5 & 37,190  & 14,376 & CNN & Accuracy
		\\
            Office-Caltech & 1,771  & 776 & AlexNet & Accuracy
		\\
		DomainNet & 6,000  & 4,573 & AlexNet & Accuracy
		\\
		BDD100K & ~60,000  & ~30,000 & YOLOv5n & mAP@0.5
		\\
		Pascal VOC & 1,464 & 1,449 & DeepLabV3 & mIoU
		\\
		MPII & 22,246 & 2,958 & PoseResNet & PCK@0.5
		\\
		\bottomrule[1pt]
		\label{tab:benchmarks}
	\end{tabular}
	\vspace{-2ex}
\end{table}

\section{Attribute Distributions of BDD100K under Different Splits}
\label{app:visual}

Since it is difficult to measure the label distribution skew for object detection datasets as each image will contain multiple objects belonging to different kinds of categories, we turn to measure the feature distribution shift across clients. BDD100K is a dataset that provides rich information about the collected images, including three attributes, i.e., weather, scene, time-of-day. It contains diverse scene types including city streets, residential areas, and highways, and diverse weather conditions at different times of the day. Therefore, those attributes can be utilized to simulate feature distribution shift by manipulating the percentage of images belonging to each specific attribute, just similar to the label-based dataset split. 

It is worth noting that each image will have all those three attributes, which means the distributions of those attributes are not independent. The first way is to choose one attribute as the main attribute (e.g., weather) for data split, where label-based split methods can be directly applied, such as IID allocation and Dirichlet-based allocation (Dir). The second way is to utilize all three attributes in a hierarchical split manner, which we call hierarchical Dirichlet-based allocation (H-Dir). More precisely, we first select an attribute as the main attribute and for each category in this attribute, then we apply the Dirichlet-based allocation to get the sample proportions across clients. Next, for each client and for the samples of each specific category in the main attribute, we repeat the Dirichlet-based allocation based on the second and third attributes. For each attribute triplet (weather, scene, time-of-day), we make the random sampling to allocate images to each client. Considering the category imbalance issue, we will use a Dirichlet concentration parameter proportional to its prior for each category in each attribute. Moreover, the calculated data amount under each triplet of attributes for each client may exceed the total amount of images belonging to that attribute triplet, where we just set as the amount of available images.  By default, we use the weather as the main attribute. 

\begin{figure*}[ht]
	\centering
	\subfigure[Dir(0.5)]{
		\includegraphics[width=0.45\columnwidth]{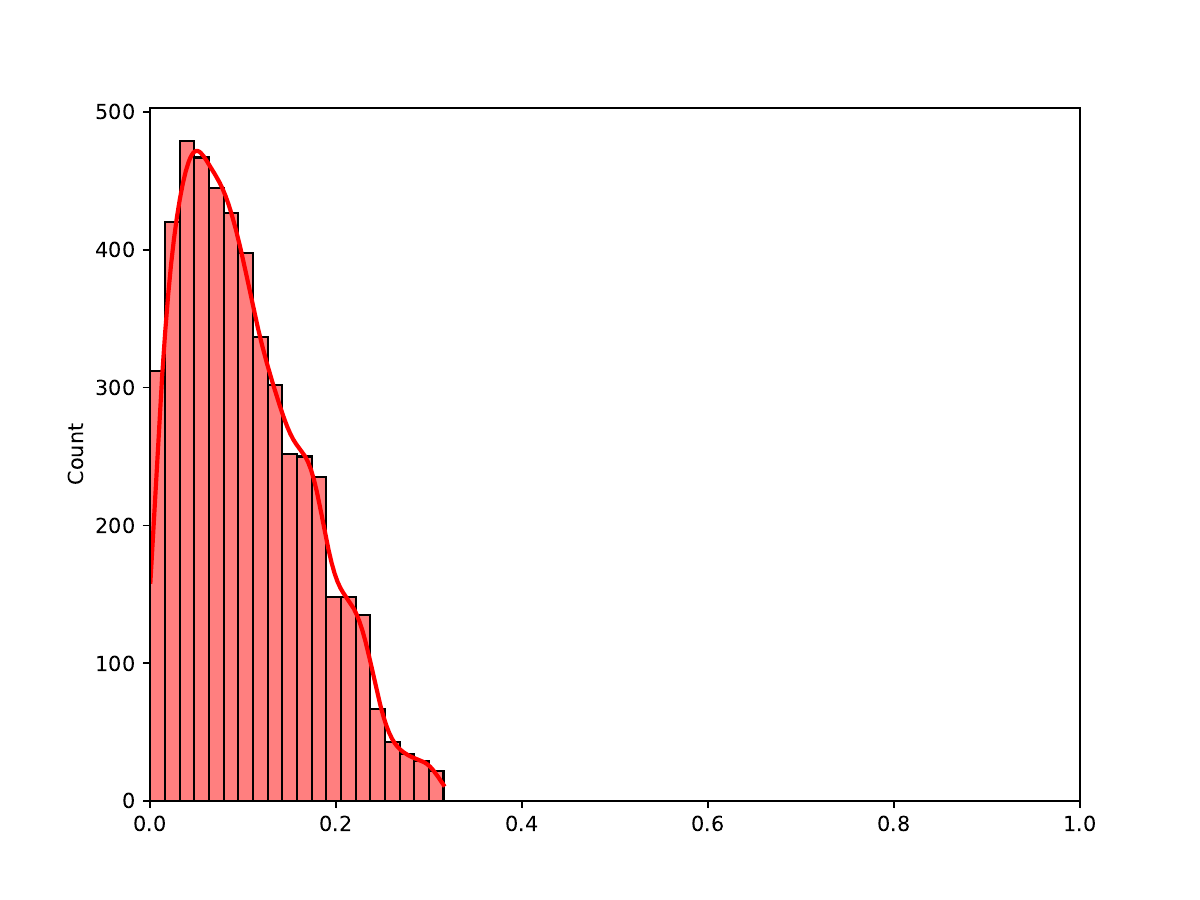}
	}
	\subfigure[H-Dir(0.5)]{
		\includegraphics[width=0.45\columnwidth]{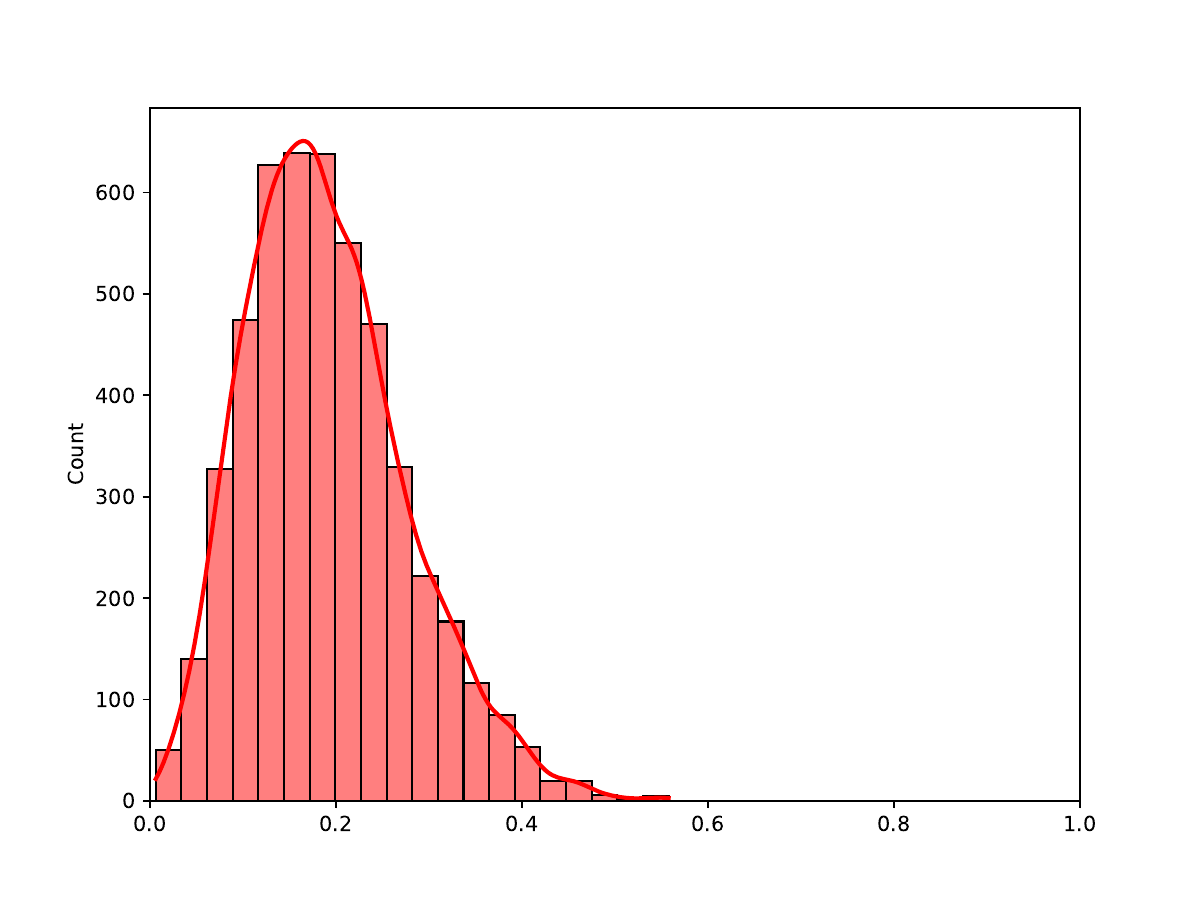}
	}
	    \vspace{-1ex}
	\caption{Pairwise Jensen–Shannon divergence of attribute distributions across clients. (a) Divergence under Dir(0.5) split; (b) Divergence under H-Dir(0.5) split. The hierarchical Dirichlet-based split will result in more divergent data distributions across clients.}
	\label{fig:bdd_divergence}
\end{figure*}

\begin{figure*}[ht]
	\centering
	\subfigure[Weather Dir(0.5)]{
		\includegraphics[width=0.8\columnwidth]{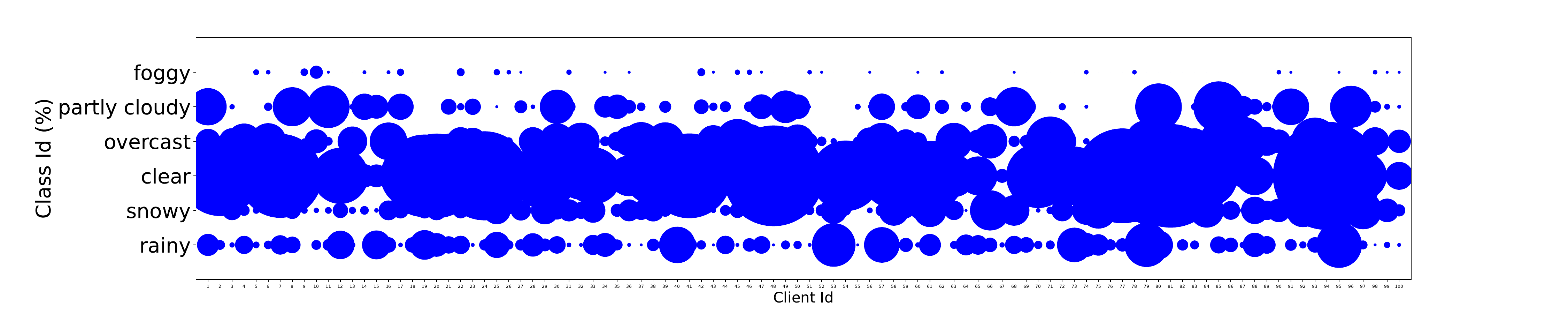}
	}
	\subfigure[Scene Dir(0.5)]{
		\includegraphics[width=0.8\columnwidth]{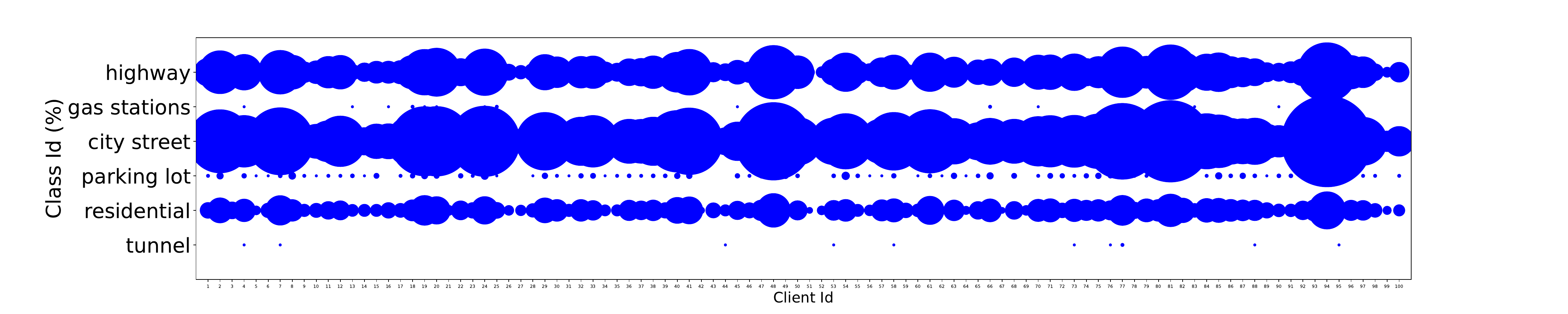}
	}
	\subfigure[Time-of-day Dir(0.5)]{
		\includegraphics[width=0.8\columnwidth]{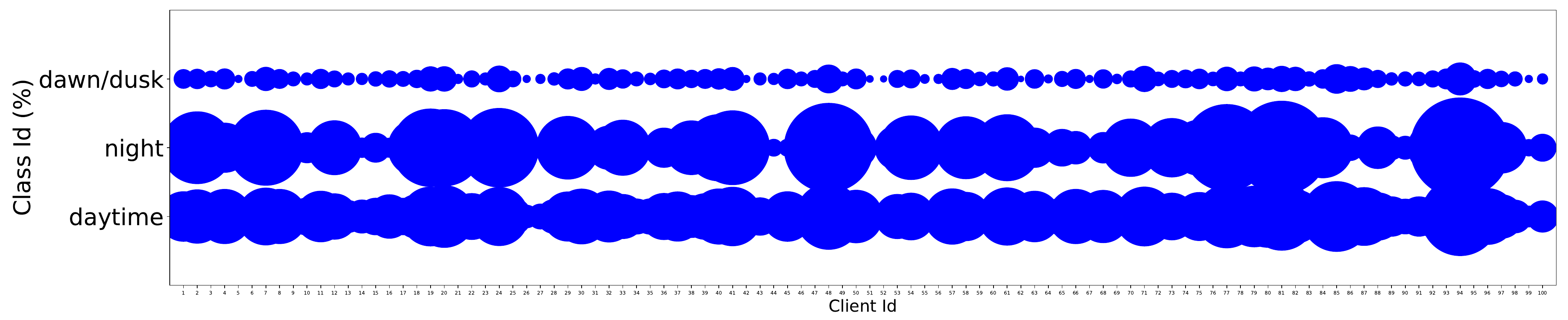}
	}

	\subfigure[Weather H-Dir(0.5)]{
		\includegraphics[width=0.8\columnwidth]{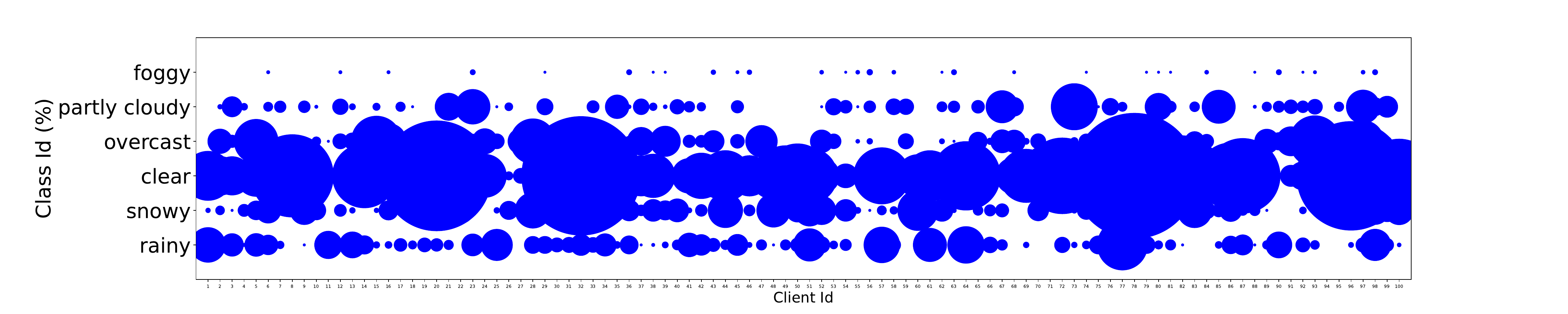}
	}
	\subfigure[Scene H-Dir(0.5)]{
		\includegraphics[width=0.8\columnwidth]{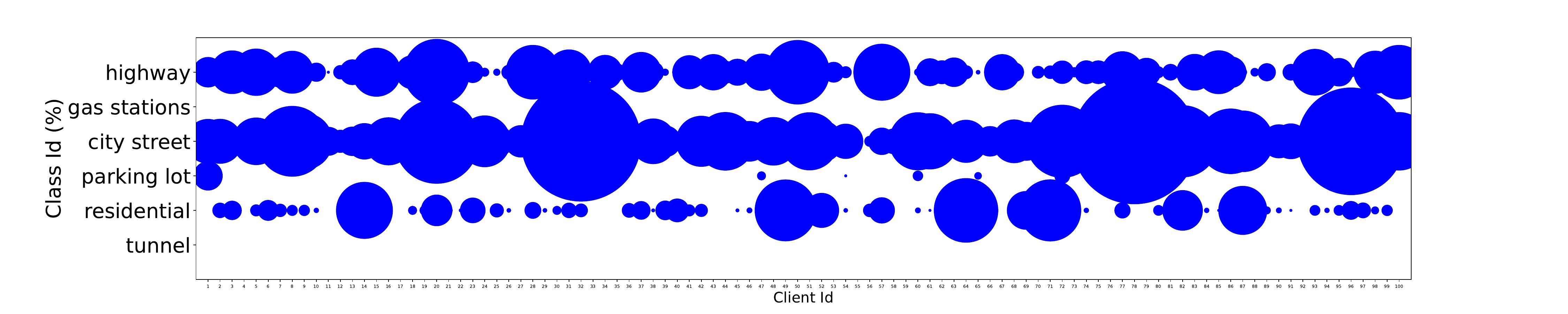}
	}
	\subfigure[Time-of-day H-Dir(0.5)]{
		\includegraphics[width=0.8\columnwidth]{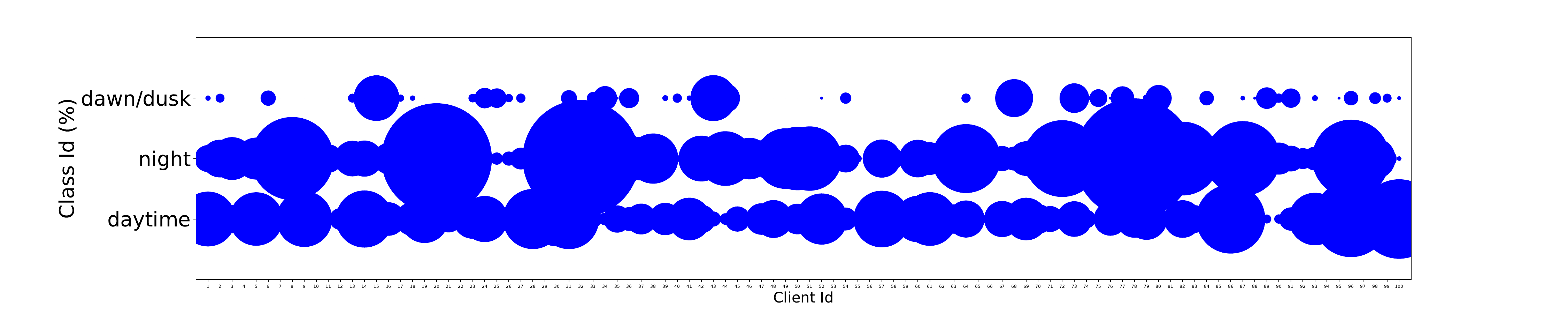}
	}
	    \vspace{-1ex}
	\caption{Attribute distributions across 100 clients under Dir (0.5) and H-Dir(0.5) data allocations. (a)-(c) are Dirichlet-based allocations only considering the weather attribute; (d)-(f) are Hierarchical Dirichlet-based allocations considering all three attributes.}
	\label{fig:bdd_dir_hdir}
\end{figure*}

In the following, we make some visualizations for training data distributions. Figure~\ref{fig:bdd_divergence} presents the statistics of pairwise Jensen–Shannon divergence that measures the difference in attribute distributions across different clients. From which we can find that the hierarchical Dirichlet-based allocation will indeed result in more divergent feature distributions across clients. We also visualize the amount of samples allocated to each client that belong to different categories in each attribute in Figure~\ref{fig:bdd_dir_hdir}. It can be found that compared with the Dirichlet-based allocation on only one attribute, the hierarchical Dirichlet-based allocation considering all three attributes can improve the imbalance level across clients for each attribute and each category.

\section{Basic Steps for Customizing FL Applications} \label{app:customize}
With the help of \name{}, it is easy to construct a new FL task with customized datasets and models. Here, we provide some general steps for customizing new FL applications. Note that not all steps are always required. For example, if the users want to only customize the model while reusing the datasets provided in \name{}, they can skip the first step of customizing the dataset. The main steps are listed as follows.

\textbf{Step 1: Customize Dataset.}

The preparation of training data is the first step for FL applications as different vision tasks may contain various kinds of data and annotation formats. Different data transformations and split strategies should also be considered. 
In the platform, we provide some basic implementations for incorporating Tensor Data, Image Data, and Torch style dataloader. For frequently used datasets in research, we directly provide the functions to load them without extra effort. In addition, we provide many examples that users can refer to for customizing the datasets easier and faster. Users can register their customized dataset into the \name{} with a simple \texttt{register\_dataset} API.


\begin{lstlisting}
import coala
# train_data, test_data are customized datasets provided the user
coala.register_dataset(train_data, test_data)
\end{lstlisting}

\textbf{Step 2: Customize Model.}

We have provided many models on \name{}, while users can also customize models for their specific applications or algorithms. They can register the models through \texttt{register\_model} API:

\begin{lstlisting}
import coala
from model import new_model
# new_model is customized by the user
coala.register_model(new_model)
\end{lstlisting}

\textbf{Step 3: Customize Client and Server Executions.}

\name{} allows users to customize client and server executions by customizing components or functions in the client and server by writing customized classes. They can just customize the parts they want and reuse the other parts of the system. For example, if users want to change the loss function, they can simply have a CustomizedClient that implements \texttt{load\_loss\_fn} with the new loss function for training. The following is the skeleton of a small subset of components for Client customization.

\begin{lstlisting}
import coala
from coala.client import BaseClient

class CustomizedClient(BaseClient):
    def __init__(*args, **kwargs):
        # initialization

    # ... many more functions ...
    
    def load_loss_fn(self, conf):
        # customized loss function implementation
        
    def train(self, conf):
        # customized client training process implementation
        
    def test(self, conf):
        # customized client evaluation process implementation
        
    def construct_upload_request(self):
        # customized content to upload to the server implementation

    # ... many more functions ...
    
coala.register_client(CustomizedClient)
\end{lstlisting}

\begin{lstlisting}
import coala
from coala.server import BaseServer

class CustomizedServer(BaseServer):
    def __init__(*args, **kwargs):
        # initialization

    # ... many more functions ...
    
    def aggregation(self):
        # customized aggregation strategy implementation

    # ... many more functions ...

coala.register_server(CustomizedServer)
\end{lstlisting}

\textbf{Step 4: Customize Configurations and Start Training.}

The configuration includes the FL training system setup, such as the number of clients, local training epochs, the data splitting method, and training hype parameters, such as local optimizer, learning rate, and local epochs. Users can simply customize the configurations without the prior steps of customizations. These configuration details can be included in a ``yaml'' file or directly defined as a dictionary in a python file, which will be merged into the default configurations.

\begin{lstlisting}[gobble=2]
  config = {
        "data": {
            "dataset": "domainnet",
            "split_type": "iid",
            "num_of_clients": 6,
        },
        "server": {
            "rounds": 100,
            "clients_per_round": 6,
        },
        "client": {
            "local_epoch": 5,
            "optimizer": {
                "type": "SGD",
                "lr": 0.01,
                "momentum": 0.9,
                "weight_decay": 0.0005,
            },
        },
        "model": "alexnet",
        "test_mode": "test_in_client",
    }
  \end{lstlisting}

\section{Comparison with Existing FL Libraries, Benchmarks, Platforms}

We have summarized and compared our proposed COALA with representative prior benchmarks on task-level in Table \ref{tab:task-level-comparison}, data level in Table \ref{tab:data-level-comparison}, and model-level in Table \ref{tab:model-level-comparison}. Our COALA has more comprehensive coverage on task level, data level, and model level compared with these prior works. The following are the compared prior works: LEAF \cite{leaf}, FedCV \cite{He21FedCV} based on FedML \cite{he2020fedml}, FedScale \cite{lai2022fedscale}, OARF \cite{hu2022oarf}, FedReID \cite{zhuang2020fedreid}, FLamby \cite{terrail2022flamby}, \cite{Xie23FederatedScope} with pFL-Bench \cite{chen2022pflbench}, Felicitas \cite{Zhang22Felicitas}, FLGO \cite{wang2023flgo}, and EasyFL \cite{Zhuang22EasyFL} \footnote{The reference the implementation from their paper and open-source codes (if available) dated the paper submission date (01/02/2024)}.

\begin{table}[!htp]\centering
    \caption{Task-level comparison of our proposed COALA and prior FL libraries, benchmarks, and platforms.}\label{tab:task-level-comparison}
    \small
    \begin{tabular}{lccccccc}\toprule
    \multirow{2}{*}{Benchmarks} &  \multirow{2}{*}{Classification} &  \multirow{2}{*}{Object Detection} & \multirow{2}{*}{Segmentation} & \multirow{2}{*}{Person ReID} &  \multirow{2}{*}{Face Recognition} &  {Pose Estimation,}\\
    &&&&&& 3D Point Cloud, etc. \\\midrule
    LEAF &$\checkmark$ &$\times$ &$\times$ &$\times$ &$\checkmark$ &$\times$ \\
    FedML (FedCV) &$\checkmark$ &$\checkmark$ &$\checkmark$ &$\times$ &$\times$ &$\times$ \\
    FedScale &$\checkmark$ &$\checkmark$ &$\times$ &$\times$ &$\times$ &$\times$ \\
    OARF &$\checkmark$ &$\times$ &$\times$ &$\times$ &$\checkmark$ &$\times$ \\
    FedReID &$\times$ &$\times$ &$\times$ &$\checkmark$ &$\times$ &$\times$ \\
    FLamby &$\checkmark$ &$\times$ &$\checkmark$ &$\times$ &$\times$ &$\times$ \\
    FederatedScope &$\checkmark$ &$\times$ &$\times$ &$\times$ &$\times$ &$\times$ \\
    Felicitas &$\checkmark$ &$\times$ &$\times$ &$\times$ &$\times$ &$\times$ \\
    FLGO &$\checkmark$ &$\checkmark$ &$\checkmark$ &$\times$ &$\times$ &$\times$ \\
    EasyFL &$\checkmark$ &$\times$ &$\times$ &$\checkmark$ &$\times$ &$\times$ \\ \midrule
    \textbf{COALA (Ours)} &$\checkmark$ &$\checkmark$ &$\checkmark$ &$\checkmark$ &$\checkmark$ &$\checkmark$ \\
    \bottomrule
    \end{tabular}
\end{table}

\begin{table}[!htp]\centering
    \caption{Data-level comparison of our proposed COALA and prior FL libraries, benchmarks, and platforms.}\label{tab:data-level-comparison}
    \small
    \begin{tabular}{lccccccc}\toprule
    \multirow{2}{*}{Benchmarks} & \multirow{2}{*}{Supervised FL} &  \multirow{2}{*}{Semi-supervised FL} & \multirow{2}{*}{Unsupervised FL} & \multirow{2}{*}{Label Shift} & \multirow{2}{*}{Domain Shift} &Continual Shift, \\
    &&&&&& Test-time Shift \\\midrule
    LEAF &$\checkmark$ &$\times$ &$\times$ &$\checkmark$ &$\times$ &$\times$ \\
    FedML (FedCV) &$\checkmark$ &$\times$ &$\times$ &$\checkmark$ &$\checkmark$ &$\times$ \\
    FedScale &$\checkmark$ &$\times$ &$\times$ &$\checkmark$ &$\checkmark$ &$\times$ \\
    OARF &$\checkmark$ &$\times$ &$\times$ &$\checkmark$ &$\checkmark$ &$\times$ \\
    FedReID &$\checkmark$ &$\times$ &$\checkmark$ &$\times$ &$\checkmark$ &$\times$ \\
    FLamby &$\checkmark$ &$\times$ &$\times$ &$\checkmark$ &$\checkmark$ &$\times$ \\
    FederatedScope &$\checkmark$ &$\times$ &$\times$ &$\checkmark$ &$\times$ &$\times$ \\
    Felicitas &$\checkmark$ &$\checkmark$ &$\times$ &$\checkmark$ &$\times$ &$\times$ \\
    FLGO &$\checkmark$ &$\times$ &$\times$ &$\checkmark$ &$\checkmark$ &$\times$ \\
    EasyFL &$\checkmark$ &$\times$ &$\checkmark$ &$\checkmark$ &$\times$ &$\times$ \\\midrule
    \textbf{COALA (Ours)} &$\checkmark$ &$\checkmark$ &$\checkmark$ &$\checkmark$ &$\checkmark$ &$\checkmark$ \\
    \bottomrule
    \end{tabular}
\end{table}

\begin{table}[!htp]\centering
    \caption{Model-level comparison of our proposed COALA and prior FL libraries, benchmarks, and platforms.}\label{tab:model-level-comparison}
    \begin{tabular}{lccccc}\toprule
    Benchmarks &Single Model &Federated Split Learning &Personalized Models &FedPEFT for FM \\\midrule
    LEAF &$\checkmark$ &$\times$ &$\times$ &$\times$ \\
    FedML (FedCV) &$\checkmark$ &$\checkmark$ &$\checkmark$ &$\times$ \\
    FedScale &$\checkmark$ &$\times$ &$\times$ &$\times$ \\
    OARF &$\checkmark$ &$\times$ &$\checkmark$ &$\times$ \\
    FedReID &$\checkmark$ &$\times$ &$\checkmark$ &$\times$ \\
    FLamby &$\checkmark$ &$\times$ &$\checkmark$ &$\times$ \\
    FederatedScope &$\checkmark$ &$\times$ &$\checkmark$ &$\times$ \\
    Felicitas &$\checkmark$ &$\times$ &$\checkmark$ &$\times$ \\
    FLGO &$\checkmark$ &$\times$ &$\checkmark$ &$\times$ \\
    EasyFL &$\checkmark$ &$\times$ &$\times$ &$\times$ \\ \midrule
    \textbf{COALA (Ours)} &$\checkmark$ &$\checkmark$ &$\checkmark$ &$\checkmark$ \\
    \bottomrule
    \end{tabular}
\end{table}

\section{System Component Overview}  \label{app:component}
The system architecture comprises both the FL server and FL client, each housing a suite of components crucial for supporting a spectrum of realistic FL scenarios.

\textit{Task Scheduler}:
The task scheduler plays a pivotal role in enabling the execution of multiple tasks by scheduling them based on the availability of resources. Tasks are queued for execution, and the resource manager collaborates to allocate resources efficiently.

\textit{Resource Manager}:
Operating in both the client and server environments, the resource manager keeps meticulous records of client and server resources, including energy consumption, computation capacity, memory, and network connectivity. It relays the availability of clients to the server and handles the allocation of computation resources for training and testing tasks. The server resource manager coordinates with the task scheduler to manage server resources, ensuring optimized execution.

\textit{Client Selector}:
This component implements algorithms for the selection of clients based on their availability. Users have the flexibility to customize client selection algorithms.

\textit{Aggregator}:
At the core of the server, the aggregator executes FL server processes, supporting a variety of aggregation methods such as FedAvg \cite{fedavg} by default, FedProx \cite{fedprox}, and FedYogi \cite{reddi2020fedyogi}. Users can use these methods directly or customize new aggregation methods.

\textit{Compute Engine}:
The compute engine in the client supports model training and evaluation, with primary compatibility for PyTorch and additional support for frameworks like TensorFlow. In the server, the compute engine is optimized for realistic FL settings, enabling further training with available public data. This is particularly beneficial for semi-supervised FL algorithms, leveraging server data for initial model training and subsequent fine-tuning with aggregated models from clients.

\textit{Data Manager and Model Manager}:
The data manager oversees data loading and partitioning for both simulation and real-world training, while the model manager facilitates model loading and customization. The \name{} platform offers a variety of pre-packaged datasets and models (refer to Table \ref{tab:overview}), and users can effortlessly extend and implement their datasets and models using these components.

\textit{Communicator}:
Responsible for managing remote communication between the server and clients, the communicator employs gRPC as the default protocol framework. It utilizes the industry-standard Protobuf for data serialization, with additional support for MQTT in cases where FL clients are deployed to Internet of Things (IoT) devices.

\textit{Tracker}:
This module collates evaluation metrics from both server and clients, capturing both system and algorithmic metrics. Users enjoy the flexibility to customize their tracking service, enabling metric storage in various formats such as disk files or through integration with external libraries or services like wandb \footnote{https://wandb.ai} or tensorboard \cite{abadi2016tensorflow-tensorboard}.

\textit{Drift Detector}:
Designed for practical scenarios involving continual learning, the drift detector continuously monitors data distribution in clients. Upon detecting drifts, it intelligently caches data to client storage and promptly notifies the server. The server, in turn, collects information from clients and initiates federated continual learning on the affected clients, ensuring adaptability to evolving data patterns.



\end{document}